\documentclass{article}

\usepackage[T1]{fontenc}
\usepackage{tgpagella}
\usepackage{graphicx,amsmath,amsfonts,amscd,amssymb,color,latexsym}
\usepackage{fullpage}
\usepackage[small,bf]{caption}
\usepackage{subfig,wrapfig}
\usepackage{bbm,bm,url}
\usepackage{microtype}
\usepackage[toc, page]{appendix}
\usepackage{algorithm,algorithmicx}
\usepackage{algpseudocode}
\usepackage{hyperref,cleveref}
\usepackage{verbatim}
\usepackage{framed}
\usepackage{comment}
\usepackage{tikz}
\usepackage{fge}
\usepackage{mathdots,mathabx}
\usepackage[margin=0.95in]{geometry}
\usepackage{pgfplots}
\usetikzlibrary{positioning}
\usetikzlibrary{decorations}
\usepackage{authblk}

\usepackage[inline]{enumitem}

\usepackage[numbers,sort&compress]{natbib}

\allowdisplaybreaks
\numberwithin{equation}{section}
\hypersetup{
    colorlinks=true,%
    citecolor=blue,%
    filecolor=blue,%
    linkcolor=blue,%
    urlcolor=blue
}

\newtheorem{theorem}{Theorem}[section]

\newtheorem{definition}[theorem]{Definition}

\def \endprf{\hfill {\vrule height6pt width6pt depth0pt}\medskip}

\newcommand{\mb}{\boldsymbol}
\newcommand{\mc}{\mathcal}
\newcommand{\mf}{\mathfrak}
\newcommand{\mr}{\mathrm}

\newcommand{\bb}{\mathbb}

\newcommand{\set}[1]{\left\{ #1 \right\}}

\newcommand{\eps}{\varepsilon}

\newcommand{\shift}[2]{s_{#2}[#1]}

\newcommand{\norm}[2]{\left\| #1 \right\|_{#2}}
\newcommand{\abs}[1]{\left| #1 \right|}
\newcommand{\innerprod}[2]{\left\langle #1,  #2 \right\rangle}

\newcommand{\proj}[2]{\mathcal{P}_{#2}\left[ #1 \right]}

\renewcommand{\eps}{\varepsilon}
\newcommand{\R}{\bb R}
\newcommand{\Cp}{\bb C}

\newcommand{\indicator}[1]{\mathbbm 1_{#1}}

\newcommand{\Brac}[1]{\left\lbrace #1 \right\rbrace}
\newcommand{\brac}[1]{\left[ #1 \right]}
\newcommand{\paren}[1]{ \left( #1 \right) }

\DeclareMathOperator{\st}{s.t.}

\pgfkeys{/pgf/decoration/.cd,
         start color/.store in =\startcolor,
         end color/.store in   =\endcolor
}

\pgfdeclaredecoration{width and color change}{initial}{
 \state{initial}[width=0pt, next state=line, persistent precomputation={%
   \pgfmathdivide{50}{\pgfdecoratedpathlength}%
   \let\increment=\pgfmathresult%
   \def\x{0}%
 }]{}
 \state{line}[width=.5pt,   persistent postcomputation={%
     \pgfmathadd{\x}{\increment}%
     \let\x=\pgfmathresult%
   }]{%
   \pgfsetlinewidth{\x/40*0.075pt+\pgflinewidth}%
   \pgfsetarrows{-}%
   \pgfpathmoveto{\pgfpointorigin}%
   \pgfpathlineto{\pgfqpoint{.75pt}{0pt}}%
   \pgfsetstrokecolor{\endcolor!\x!\startcolor}%
   \pgfusepath{stroke}%
 }
 \state{final}{%
   \pgfsetlinewidth{\pgflinewidth}%
   \pgfpathmoveto{\pgfpointorigin}%
   \color{\endcolor!\x!\startcolor}%
   \pgfusepath{stroke}%
 }
}

\pgfdeclaredecoration{width and color change legend}{initial}{
 \state{initial}[width=0pt, next state=line, persistent precomputation={%
   \pgfmathdivide{100}{\pgfdecoratedpathlength}%
   \let\increment=\pgfmathresult%
   \def\x{0}%
 }]{}
 \state{line}[width=.5pt,   persistent postcomputation={%
     \pgfmathadd{\x}{\increment}%
     \let\x=\pgfmathresult%
   }]{%
   \pgfsetlinewidth{\x/40*0.075pt+\pgflinewidth}%
   \pgfsetarrows{-}%
   \pgfpathmoveto{\pgfpointorigin}%
   \pgfpathlineto{\pgfqpoint{.75pt}{0pt}}%
   \pgfsetstrokecolor{\endcolor!\x!\startcolor}%
   \pgfusepath{stroke}%
 }
 \state{final}{%
   \pgfsetlinewidth{\pgflinewidth}%
   \pgfpathmoveto{\pgfpointorigin}%
   \color{\endcolor!\x!\startcolor}%
   \pgfusepath{stroke}%
 }
}

\newcommand{\NC}{$\mc {NC}$}

\usepackage{tikz-3dplot}
\usepackage{pgfplots}
\usepgfplotslibrary{patchplots}
\usetikzlibrary{patterns, positioning, arrows}
\pgfplotsset{compat=1.15}

\newcommand{\reals}{\bb R}

\begin{document}

\title{From Symmetry to Geometry: Tractable Nonconvex Problems}

\author[$\sharp$]{Yuqian Zhang}
\author[$\Diamond$]{Qing Qu}
\author[$\dagger$,$\ddagger$]{John Wright}

\affil[$\sharp$]{Department of Electrical \& Computer Engineering, Rutgers University}
\affil[$\Diamond$]{Department of Electrical Engineering and Computer Science, University of Michigan}
\affil[$\dagger$]{Department of Electrical Engineering and Data Science Institute, Columbia University}
\affil[$\ddagger$]{Department of Applied Physics and Applied Mathematics, Columbia University }
\date{}
\maketitle

\begin{abstract}
{\normalsize
As science and engineering have become increasingly data-driven, the role of optimization has expanded to touch almost every stage of the data analysis pipeline, from signal and data acquisition to modeling and prediction. The optimization problems encountered in practice are often nonconvex. While challenges vary from problem to problem, one common source of nonconvexity is {\em nonlinearity} in the data or measurement model. Nonlinear models often exhibit {\em symmetries}, creating complicated, nonconvex objective landscapes, with multiple equivalent solutions. Nevertheless, simple methods (e.g., gradient descent) often perform surprisingly well in practice.

The goal of this survey is to highlight a class of tractable nonconvex problems, which can be understood through the lens of symmetries. These problems exhibit a characteristic geometric structure: local minimizers are symmetric copies of a single ``ground truth'' solution, while other critical points occur at balanced superpositions of symmetric copies of the ground truth, and exhibit negative curvature in directions that break the symmetry. This structure enables efficient methods to obtain global minimizers. We discuss examples of this phenomenon arising from a wide range of problems in imaging, signal processing, and data analysis. We highlight the key role of symmetry in shaping the objective landscape and discuss the different roles of rotational and discrete symmetries. This area is rich with observed phenomena and open problems; we close by highlighting directions for future research.
}
\end{abstract}


\tableofcontents

%
%
%
%
%
%

\section{Introduction}
\label{sec:intro}

\definecolor{royalpurple}{rgb}{0.47, 0.32, 0.66}
\definecolor{lava}{rgb}{0.81, 0.06, 0.13}
\definecolor{darkblue}{rgb}{0, 0, 0.66}

As engineering and the sciences become increasingly data and computation driven, the role of optimization has expanded to touch almost every stage of the data analysis pipeline, from the signal and data acquisition to modeling and prediction. While the challenges in computing with physical data are many and varied, basic recurring issues arise from {\em nonlinearities} at different stages of this pipeline:

\begin{quote}
\begin{itemize}[leftmargin=*]
	\item {\bf \em Nonlinear measurements} \em are ubiquitous in imaging, optics, and astronomy. A canonical example is sensing magnitude measurements, which arises when it is easy to measure the (Fourier) modulus of a complex signal, but hard to measure the phase. For example, we might measure the Fourier magnitude of a complex signal $\mb x\in\Cp^n$ \cite{patterson1934fourier, patterson1944ambiguities, shechtman2015phase, jaganathan2017phase}
\begin{align}\label{eqn:phase-measurement}
\underset{\text{\color{lava}\bf observation}}{\mb y} = \abs{\mc F \left(\underset{\text{\color{lava}\bf unknown signal}}{\mb x} \right)} \in\R^m.
\end{align}
Here, $\mc F(\cdot)$ denotes the Fourier transform, $\mb x$ represents a signal or image of interest, and the goal is to reconstruct $\mb x$ from the nonlinear measurements $\mb y$.

\item {\bf \em Nonlinear models} are often well-suited to express the variability of real datasets. For example, observations in microscopy, neuroscience, and astronomy can often be approximated as sparse superpositions of basic motifs. We can cast the problem of finding these motifs as one of seeking a representation of the form
\begin{align}
\underset{\text{\color{lava}\bf data}}{\mb Y} = \underset{\text{\color{lava}\bf motifs}}{ \mb A} \;\; \underset{\text{\color{lava}\bf sparse coefficients}}{\mb X.} 
\end{align}
Here, the columns of $\mb Y \in \bb R^{m \times p}$ are observed data vectors, the columns of $\mb A \in \bb R^{m \times n}$ are basic motifs, and $\mb X \in \bb R^{n \times p}$ is a sparse matrix of coefficients that expresses each observed data point as a superposition of motifs. This is sometimes called a {\em sparse dictionary model}. A typical goal is to infer both $\mb A$ and $\mb X$ from observed data. Because both $\mb A$ and $\mb X$ are unknown, this model should be considered nonlinear (strictly, bilinear). Natural images may have even more variability, which is better modeled by hierarchical models (convolutional neural networks) with more complicated nonlinearities \cite{lecun1995convolutional, goodfellow2014generative,goodfellow2016deep}. 
\end{itemize}
\end{quote}

\paragraph{Nonlinearity, Symmetry, and Nonconvexity} In the examples described above, nonlinearities are not just a nuisance: they have strong implications on the sense in which we can hope to solve these problems, and, as we shall see in this paper, on our ability to efficiently compute solutions. Both above models exhibit {\em symmetries}. The model $\mb y = \abs{ \mc F(\mb x) }$ in \eqref{eqn:phase-measurement} exhibits a {\em phase symmetry}: both $\mb x$ and $\mb x e^{i \phi}$ (for any $\phi \in [0, 2\pi)$) produce the same observation $\mb y$. The sparse dictionary model $\mb Y = \mb A \mb X$ exhibits a {\em permutation symmetry}: for any permutation $\mb \Pi$, $(\mb A,\mb X)$ and $(\mb A \mb \Pi, \mb \Pi^* \mb X)$ produce the same observation $\mb Y$.\footnote{Here, and throughout the paper, the notation $\mb M^*$ denotes the complex conjugate transpose of a matrix $\mb M$. If $\mb M$ is real-valued, this is simply the matrix transpose.} 

In either case, we can only hope to recover the physical ground-truth up to these basic symmetries. A typical computational approach is to formulate it as an optimization problem
\begin{equation} \label{eqn:general-opt-problem}
\min_{\mb z} \, \varphi(\mb z),
\end{equation} 
and attempt to solve it with iterative methods such as gradient descent \cite{cauchy1847methode}. Here, $\mb z$ represents the signal or model to be recovered -- for example, in phase retrieval, $\mb z = \mb x$, while in dictionary learning the optimization variable $\mb z$ is the pair $(\mb A,\mb X)$. Typically, $\varphi(\cdot)$ measures quality of fit to observed data and the extent to which the solution satisfies assumptions such as sparsity. As we shall see, most natural choices of $\varphi(\cdot)$ inherit the symmetries of the data generation model: e.g., for phase recovery, we have $\varphi( e^{i\theta } \mb x ) = \varphi(\mb x)$, while for dictionary learning, $\varphi( (\mb A, \mb X) ) = \varphi( (\mb A \mb \Pi, \mb \Pi^* \mb X ))$: {\em symmetries of the observation models become symmetries of the optimization problem}. 

If we are judicious in our choice of $\varphi(\cdot)$, we can hope that the true $\mb x$ is a (near) global minimizer; our task becomes solving the optimization problem \eqref{eqn:general-opt-problem} to the global optimality. In contrast to certain applications of numerical optimization (e.g., in finance, logistics, deep learning, etc.), we care not just about decreasing the objective function, but about obtaining the physical ground truth. As such, we are forced to care not just about ensuring that our algorithms converge, but that they converge to the global minimizers. In applied optimization, a time-honored approach for guaranteeing global optimality is to seek formulations that are convex. The global minimizers of a convex function form a convex set. Moreover, every local minimizer (indeed, every critical point) of a convex function is global. As a result, many convex problems can be efficiently solved to global optimality by local methods. This makes the area of convex analysis and optimization a model for how geometric understanding can support practical computations. 

%
%
%

Unfortunately, as alluded to above, the symmetric problems we encounter in statistics, signal processing, and related areas are typically nonconvex \cite{sun2015nonconvex,jain2017non,chi2019nonconvex,sun2019link}, and so we need to look for other geometric principles that will enable us to guarantee high-quality solutions. Indeed, these problems exhibit multiple global minimizers, which may be disjoint (due to permutation symmetry) or may reside on a nonconvex set (due to rotation or phase symmetry). Any optimization formulation that inherits these symmetries will be nonconvex.\footnote{{\bf Disclaimer}: Not {\em every} symmetric problem is nonconvex. Indeed, the objective function $\varphi(\mb z) = \tfrac{1}{2} \| \mb z \|_2^2$ is rotationally symmetric $\varphi(\mb R \mb z) = \varphi(\mb z)$ for all $\mb R \in O(n)$, $\mb z \in \bb R^n$ and convex. It is easy to construct additional examples of this type. However, the symmetric problems encountered in statistics, signal processing, and related areas are typically nonconvex; moreover, their nonconvexity can be directly attributed to symmetry.}

\begin{figure}[h]
\centerline{
\begin{tikzpicture}
\node at (-2.5,0) {\includegraphics[width=2in]{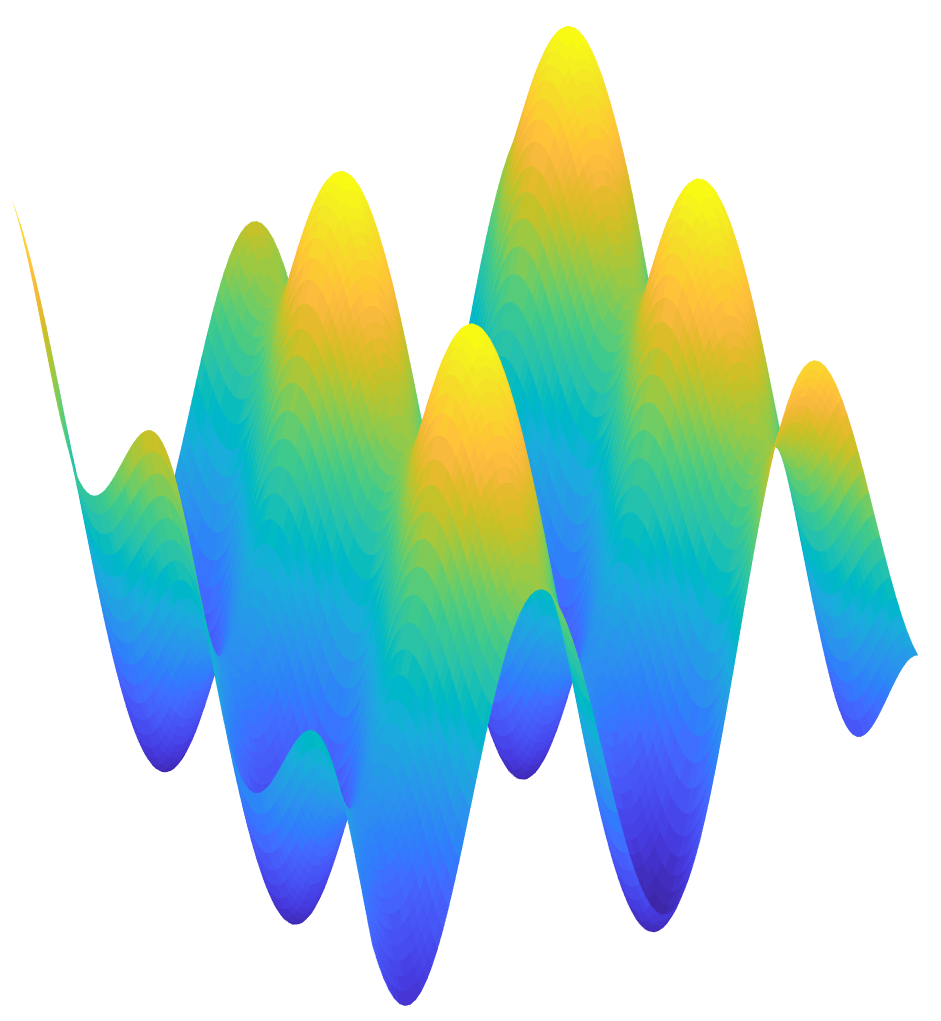}};
\node at (-2.5,-3) {\bf Spurious local minimizers};
\node at (2.5,0) {\includegraphics[width=2in]{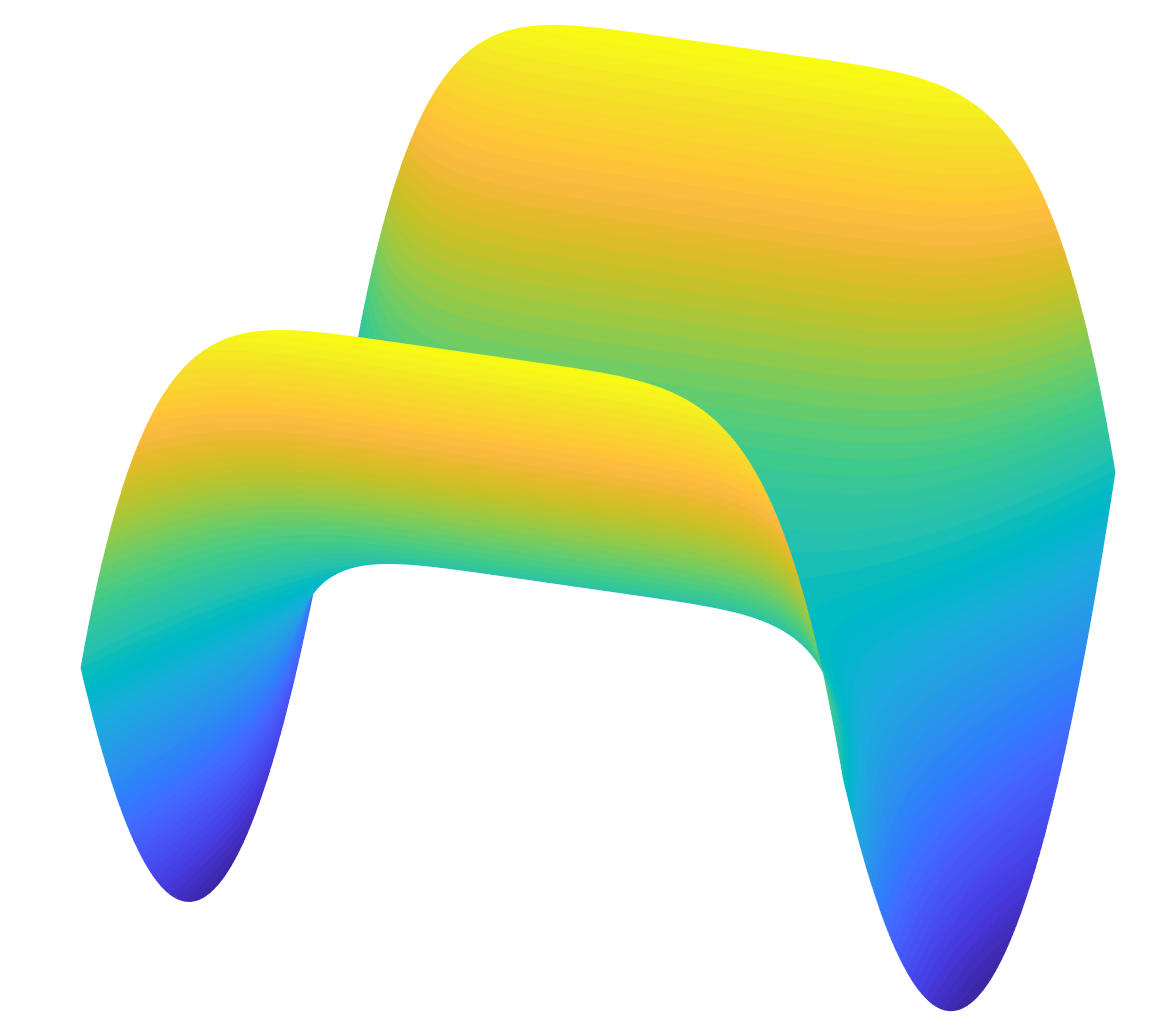}};
\node at (2.5,-3) {\bf Flat saddle points};
\end{tikzpicture}
}
\caption{{\bf Two geometric obstructions} to nonconvex optimization. Local methods can become trapped near local minimizers (left) or stagnate near flat saddle points (right).} 
\label{fig:ncvx-two-obstructions}
\end{figure}

\paragraph{Worst Case Obstructions to Nonconvex Optimization} This observation might suggest a certain pessimism: {\em nonconvex optimization is impossible in general.} There are simple classes of nonconvex problems (e.g., in polynomial optimization) that are NP-hard \cite{murty1987some}. At a more intuitive level, there are two geometric obstructions to solving nonconvex problems globally. First, nonconvex problems can exhibit {\em spurious local minimizers}, i.e. local minimizers that are not global (see \Cref{fig:ncvx-two-obstructions} (left) for an illustration). Local descent methods can get trapped; finding the global optimum is hard in general. Perhaps surprisingly, even finding a {\em local} minimizer is NP-hard in general \cite{murty1987some,nesterov2000squared}. \Cref{fig:ncvx-two-obstructions} (right) illustrates the challenge: it is possible to construct objective functions that are so flat that it is impossible to efficiently determine a descent direction.

\begin{figure}[h]
\centerline{
\begin{tikzpicture}
\node at (-3,.85) {\includegraphics[width=1.25in]{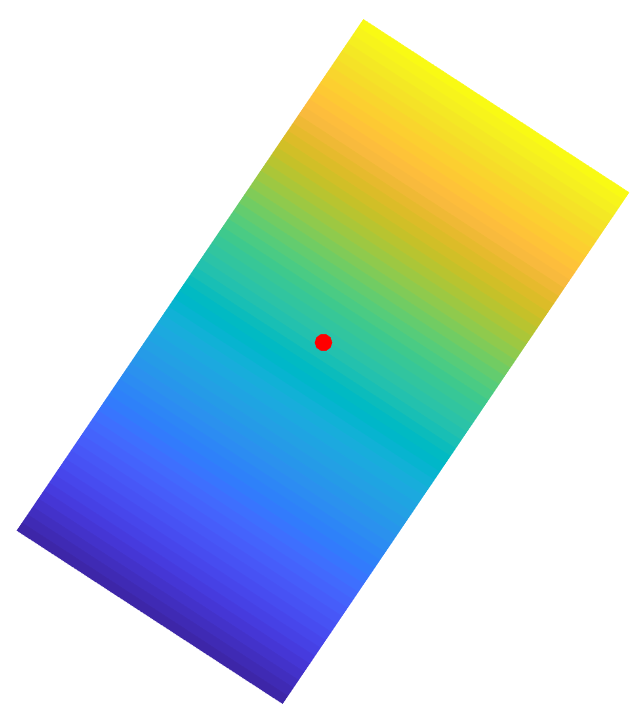}};
\node at (-3,-1.25) {\bf Noncritical Point ($\nabla \varphi \ne \mb 0$)};
\node at (4,-1.25) {\bf Critical Points ($\nabla \varphi = \mb 0$)};
\node at (.65,1.85) {\includegraphics[width=1.15in]{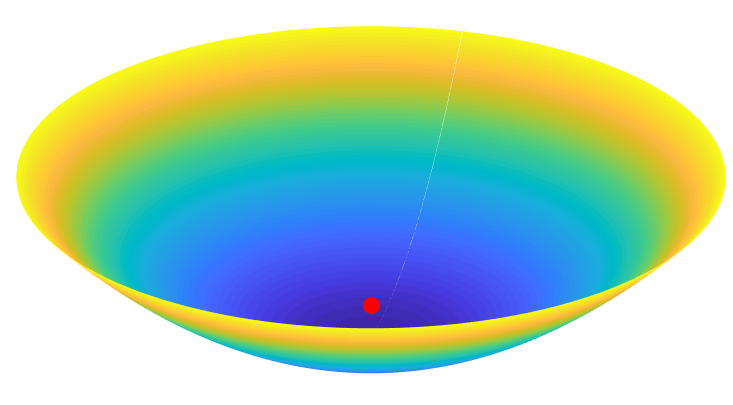}};
\node at (.65,0) {$\nabla^2 \varphi \succ \mb 0$};
\node at (.65,.5) {Minimizer};
\node at (4,1.85) {\includegraphics[width=1.15in]{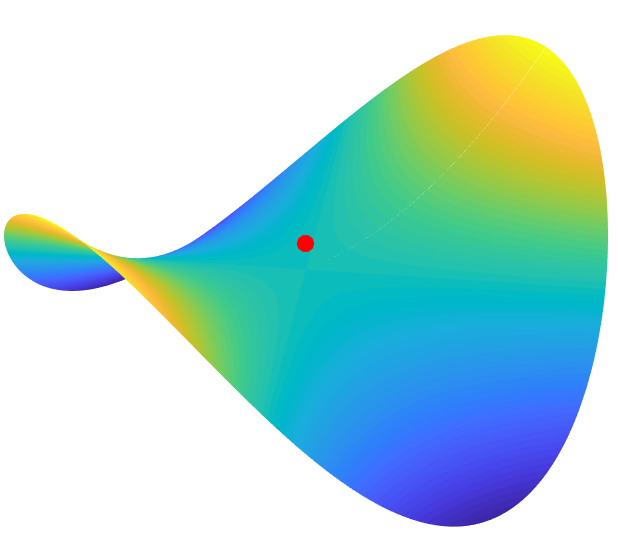}};
\node at (4,0) {$\lambda_{\min} \nabla^2 \varphi < 0$};
\node at (4,-.5) {$\lambda_{\max} \nabla^2 \varphi > 0$};
\node at (4,.5) {Saddle};
\node at (7.35,0) {$\nabla^2 \varphi \prec \mb 0$};
\node at (7.35,.5) {Maximizer};
\node at (7.35,1.85) {\includegraphics[width=1.15in]{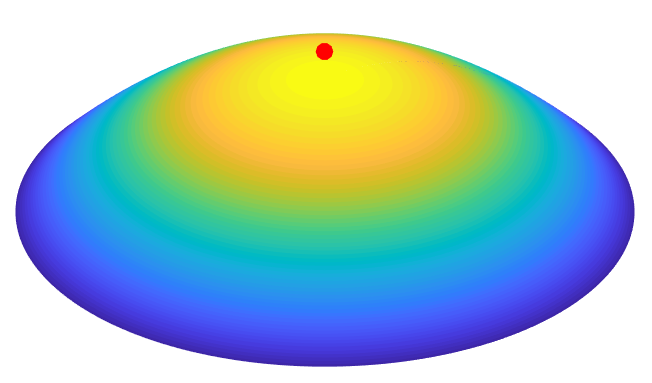}};
\end{tikzpicture}}
\caption{{\bf Calculus and the {\em local} geometry of optimization.} The {\em gradient} $\nabla \varphi$ captures the slope of the function $\varphi$. At {\em critical points} $\bar{\mb z}$, $\nabla \varphi(\bar{\mb z}) = \mb 0$. The type of critical point (minimizer, maximizer, saddle) can often be determined from the curvature of $\varphi$ at $\bar{\mb z}$, which is captured by the {\em Hessian} $\nabla^2 \varphi(\bar{\mb z})$.} 
\label{fig:grad_hess}
\end{figure}
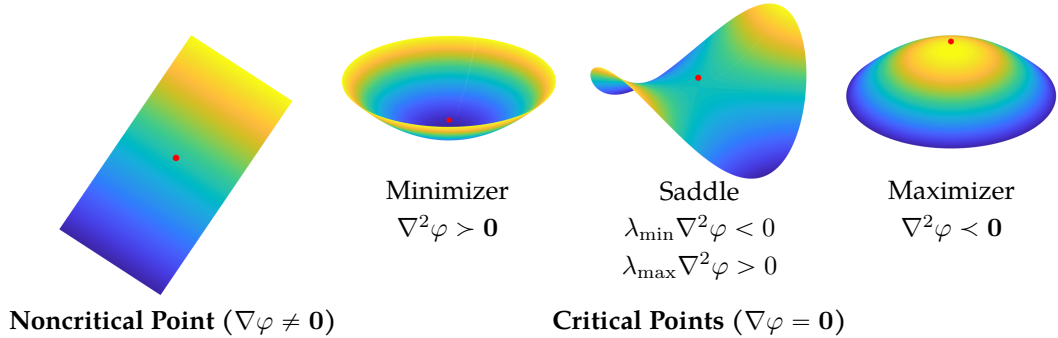

\paragraph{Calculus and the Local Geometry of Optimization} 

Because of these worst case obstructions, the classical literature on efficient\footnote{Of course, it is also possible to find global optima under minimal assumptions by exhaustively exploring the space of optimization, e.g., by discretization \cite{erdogdu2018global} or by random search \cite{hajela1990genetic,burke2005robust}. The worst case obstructions described above still rear their heads, in the form of search times that are exponential in dimension.}  nonconvex optimization has focused on guaranteeing 
\begin{quote}
\begin{enumerate}
	\item[(i)] convergence to some critical point ($\bar{\mb z}$ such that $\nabla \varphi(\bar{\mb z}) = \mb 0$), 
	\item[(ii)] or convergence to some local minimizers, for functions $\varphi(\cdot)$ which are not too flat.
\end{enumerate}
\end{quote}
The curvature of a smooth function $\varphi(\cdot)$ around a critical point $\bar{\mb z}$ can be studied through the Hessian $\nabla^2 \varphi(\bar{\mb z})$. If $\nabla^2 \varphi(\bar{\mb z})$ is nonsingular, the signs of its eigenvalues completely determine whether $\bar{\mb z}$ is a minimizer, maximizer, or saddle point -- see \Cref{fig:grad_hess} (right). In particular, if $\bar{\mb z}$ is a saddle point or a minimizer, there is a direction of negative curvature -- a direction along which the second derivative is negative. This information can be used to escape saddles and converge to a local minimizer, either explicitly (using the Hessian \cite{absil2007trust,conn2000trust,sun2017complete2}) or implicitly (using gradient information \cite{lee2016gradient,ge2015escaping,jin2021nonconvex}). There are a variety of iterative methods that trade-off in various ways between the amount of computation used to determine a good direction of negative curvature at a given iteration and the number of iterations required to converge \cite{goldfarb1980curvilinear,conn2000trust,nesterov2006cubic,lee2016gradient,jin2017accelerated,lee2019first}. However, the high-level message of these methods is consistent: if all critical points are nondegenerate,\footnote{In the language of topology, if the function $\varphi$ is Morse \cite{milnor1969morse,bott1982lectures}.} we can escape them and efficiently converge to a local minimizer. In fact, slightly less is required: it is enough that every non-minimizing critical point have a direction of strict negative curvature \cite{jin2017escape,jin2017accelerated,lee2017first,lee2016gradient}.\footnote{In the recent literature, this is called a ``strict saddle'' property \cite{ge2015escaping,sun2015nonconvex}. Concrete rates of convergence are typically stated in terms of quantitative versions of this property, which explicitly control the size of the gradient and the smallest eigenvalue of the Hessian uniformly over the domain of optimization.}

Results of this nature control the worst-case behavior of optimization methods over very broad classes of problems. In such a general setting, it is impossible to provide strong guarantees on {\em what} local minimizer that optimization methods converge to, and whether that minimizer is global. Nevertheless, it is difficult to overstate the impact of this kind of thinking for stimulating the development of useful methods and elucidating their properties. Moreover, optimization methods developed to guarantee good worst-case performance often outperform their worst-case guarantees on practical problem instances -- we witnessed longstanding ``folk theorems'' on the ease of optimizing neural networks \cite{choromanska2014loss,kawaguchi2016deep,soltanolkotabi2018theoretical,allen2019convergence,du2019gradient,sun2019optimization}, solving problems in quantum mechanics \cite{kyrillidis2018provable,sheldon2018taming,hu2019brief} or clustering separated data \cite{qian2019global,kwon2019global,qian2020structures,wang2020efficient}. Delineating problem classes that capture the difficulty (or ease!) of naturally occurring nonconvex optimization problems is a pressing challenge for the mathematics of data science \cite{sun2015nonconvex,jain2017non,chi2019nonconvex,sun2019link}.

\begin{figure}[h]
\centerline{
\begin{tikzpicture}
\node at (-3.5,0) {\includegraphics[width=2.75in]{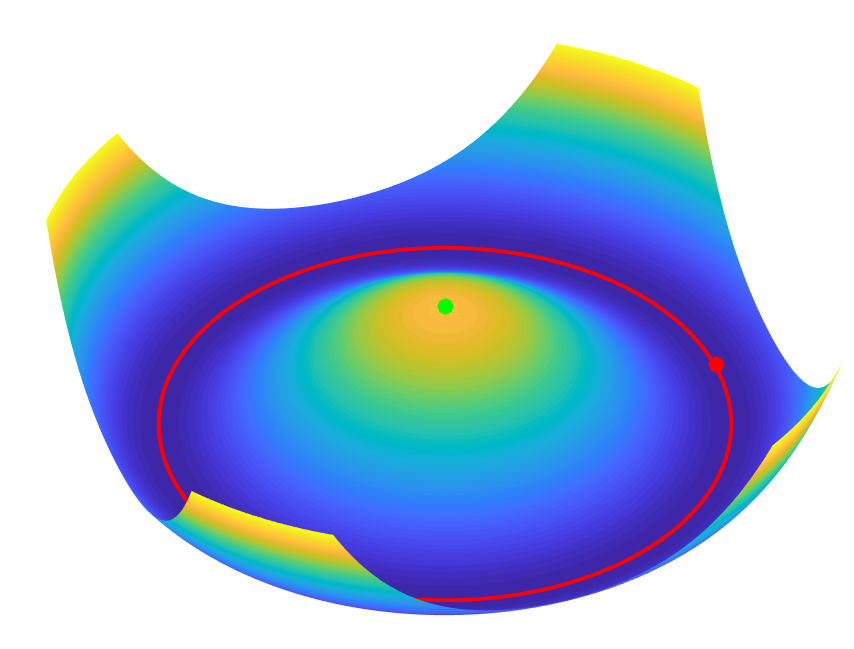}};
\node at (-3.5,-3) {\bf Rotational symmetry};
\node at (3.5,0) {\includegraphics[width=2.5in]{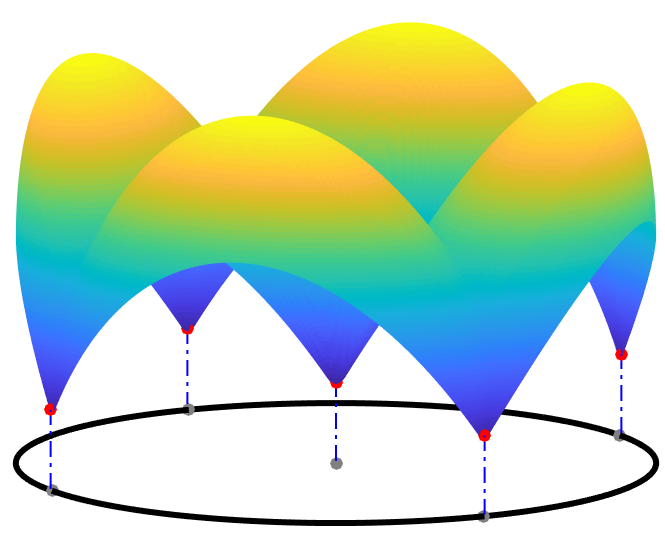}};
\node at (3.5,-3) {\bf Discrete symmetry};
\end{tikzpicture}
}
\caption{{\bf Symmetry and the {\em global} geometry of optimization.} Model problems with continuous (left) and discrete (right) symmetry. For these particular problems, and a few others we will survey, every local minimizer is global.} 
\label{fig:ncvx_examples}
\end{figure}

\paragraph{Symmetry and the Global Geometry of Optimization?} 

The goal of this survey paper is to highlight a particular family of easy nonconvex problems which, under certain hypotheses, can be solved globally with efficient optimization methods. This family includes a number of contemporary problems in signal processing, data analysis, and related fields \cite{sun2015nonconvex,jain2017non,chi2019nonconvex,sun2019link}. The most important high-level property of these problems is that they are {\em symmetric} -- in slightly more formal mathematical language:
\begin{definition}[Symmetric Function] Let $\bb G$ be a group acting on $\bb R^n$. A function $\varphi : \bb R^n \to \bb R^{n'}$ is $\bb G$-symmetric if for all $\mb z \in \reals^n$, $g \in \bb G$, $\varphi(g \cdot \mb z) = \varphi(\mb z)$. 
\end{definition}
As argued above, symmetry forces us to grapple with the properties of nonconvex functions. On the other hand, the particular symmetric nonconvex functions encountered in practice are often quite benign.  \Cref{fig:ncvx_examples} shows two examples -- one with rotational symmetry (e.g., $\bb G$ is an orthogonal group) and one with discrete symmetry (e.g., $\bb G$ is a discrete group, such as signed permutations). We will develop these examples in more mathematical detail below. For now, we simply observe that these two instances do not exhibit spurious local minimizers or flat saddle points. The absence of these worst-case obstructions can be attributed to benign symmetry structures. In slogan form, we shall see that:

\begin{quote}
\begin{enumerate}
\item[\bf Slogan I:] {\em The (only!) local minimizers are symmetric versions of the ground truth.}
\item[\bf Slogan II:] {\em There is negative curvature in directions that break symmetry.}
\end{enumerate}
\end{quote}

When these two slogans are in force, efficient (local) optimization methods produce global minimizers. Moreover, symmetry constrains the global layout of the critical points, leading to an additional structure that facilitates efficient optimization. We will show examples where the saddle points of symmetric problems  ``cascade'', with negative curvature directions feeding into other negative curvature directions subsequently, a benign property which appears to prevent first-order methods from stagnating \cite{gilboa2018efficient}. 


Before we embark, a few disclaimers are in order. First, Slogans I and II are only slogans. As we shall see, they have been established rigorously for specific problems under specific (restrictive) technical hypotheses. We hope to convey a sense of the beauty and robustness of certain observed phenomena in optimization, while also making clear that the existing mathematics supporting these claims is in places ugly and brittle. There is a need for more unified analyses and better technical tools. We highlight some potential avenues for this in \Cref{sec:discussion}. The second, more fundamental, the disclaimer is that not all symmetric problems have benign global geometry. It is easy to construct counterexamples. Nevertheless, as we will see, symmetry provides a lens through which one can understand the geometric properties that enable efficient optimization for a particular family of problems. Moreover, when we study these problems through their symmetries, common structures and intuitions emerge: problems with similar symmetries exhibit similar geometric properties.


\begin{figure}[t]
\centerline{
\begin{tikzpicture}
\draw[very thick] (-3.25,-7.35) -- (13.5,-7.35) -- (13.5,.15) -- (-3.25,.15) -- cycle;
\node at (5.1,-.5) {\bf \Large Nonconvex Problems with Rotational Symmetries};
\node (P) at (0,-1) {\footnotesize };
\node [right=-.75 of P.center, anchor=north] (PS) {\bf Eigenspace Computation};
\node [below = 2.5 of PS.center, anchor = center] {\includegraphics[width=1.3in]{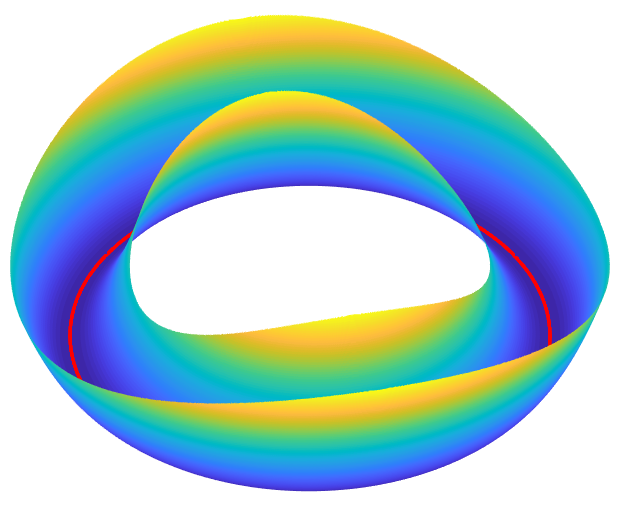}};
\node [below = 4.55 of PS.center, anchor = center] {\footnotesize $\min_{\mb X^* \mb X = \mb I} -\tfrac{1}{2} \mathrm{trace}\left[ \mb X^* \mb A \mb X \right]$.};
\node [below = .50 of PS.north, anchor = north] {\footnotesize \em Compute the principal subspace}; 
\node [below = .80 of PS.north, anchor = north] {\footnotesize \em  of a symmetric matrix.};
\node [below = 5.25 of PS.center, anchor = center] {\footnotesize \bf \em Symmetry: $\mb X \mapsto \mb X \mb R$};
\node [below = 5.65 of PS.center, anchor = center] {\footnotesize $\bb G = O(r)$};
\node [right=4.75 of P.center, anchor=north] (GPR) {\bf Generalized Phase Retrieval};
\node [below = 2.5 of GPR.center, anchor = center] {\includegraphics[width=1.75in]{figures/gpr_1c_labeled.png}};
\node [below = 4.55 of GPR.center, anchor = center] {\footnotesize $\min_{\mb x} \tfrac{1}{2} \| \mb y^{2} - | \mb A \mb x |^2 \|^2_2$.};
\node [below = .50 of GPR.north, anchor = north] {\footnotesize \em Recover a complex vector $\mb x_0$ from}; 
\node [below = .80 of GPR.north, anchor = north] {\footnotesize \em  magnitude measurements $\mb y = |\mb A \mb x_0|$.};
\node [below = 5.25 of GPR.center, anchor = center] {\footnotesize \bf \em Symmetry: $\mb x \mapsto \mb x e^{\mathfrak{i} \phi}$};
\node [below = 5.65 of GPR.center, anchor = center] {\footnotesize $\bb G =\bb S^1 \cong O(2)$};
\node [right=10.75 of P.center, anchor= north] (MR) {\bf Matrix Recovery};
\node [below = 2.5 of MR.center, anchor = center] {\includegraphics[width=1.75in]{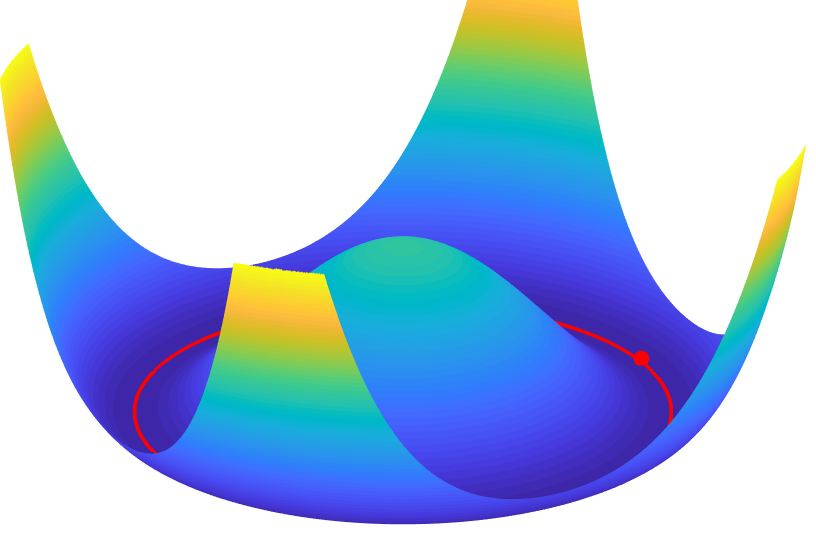}};
\node [below = .47 of MR.north, anchor = north] {\footnotesize \em Recover a low-rank matrix $\mb X = \mb U \mb V^*$}; 
\node [below = .80 of MR.north, anchor = north] {\footnotesize \em from incomplete/corrupted observations};
\node [below = 4.55 of MR.center, anchor = center] {\footnotesize   $\min_{\, \mb U,\mb V} \mc L( \mb Y -\mc A[ \mb U \mb V^* ])  \;+\; \rho(\mb U,\mb V)$.};
\node [below = 5.25 of MR.center, anchor = center] {\footnotesize \bf \em Symmetry: $(\mb U,\mb V) \mapsto (\mb U\mb \Gamma, \mb V \mb \Gamma^{-*})$};
\node [below = 5.65 of MR.center, anchor = center] {\footnotesize $\bb G = \mr{GL}(r)$ or $\bb G = O(r)$};
\end{tikzpicture}
}
    \caption{Three examples of nonconvex optimization problems with rotational symmetries (Section \ref{sec:rotational}). Each of these three tasks can be reduced to optimization problems in various ways; for each, we give a representative formulation and discuss its symmetries.} \label{fig:rotational-symmetries}
\end{figure} 


\subsection*{A Taxonomy of Symmetric Nonconvex Problems}

\begin{figure}[t]
\centerline{
\begin{tikzpicture}
\draw[very thick] (-4,-13.75) -- (9.25,-13.75) -- (9.25,.15) -- (-4,.15) -- cycle;
\node at (2.45,-.5) {\bf \Large Nonconvex Problems with Discrete Symmetries};
\node (P) at (0,-1) {\footnotesize };
\node [right=-.75 of P.center, anchor=north] (EV) {\bf Eigenvector Computation};
\node [below = 2.5 of EV.center, anchor = center] {\includegraphics[width=1.15in]{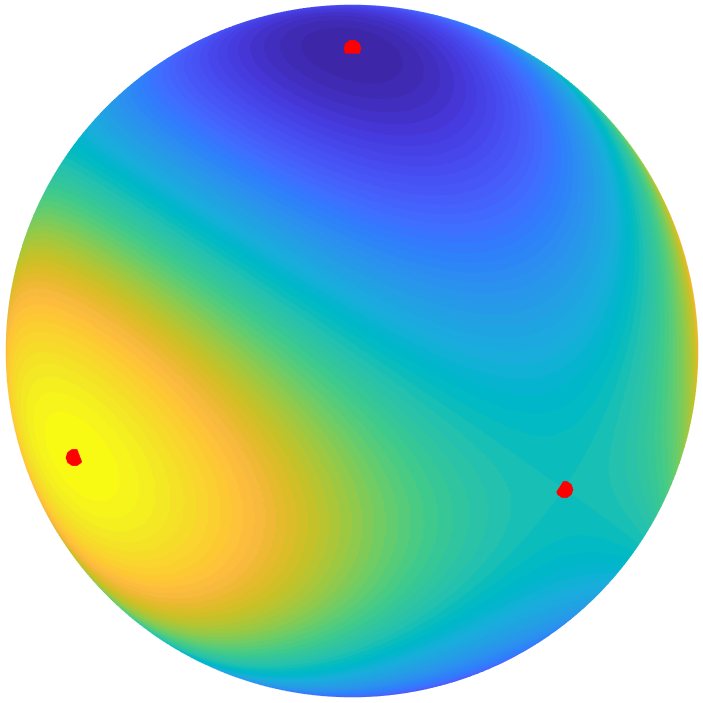}};
\node [below = 4.55 of EV.center, anchor = center] {\footnotesize $\max_{\mb x \in \bb S^{n-1}} \tfrac{1}{2} \mb x^* \mb A \mb x$.};
\node [below = .50 of EV.north, anchor = north] {\footnotesize \em Maximize a quadratic form}; 
\node [below = .80 of EV.north, anchor = north] {\footnotesize \em  over the sphere.};
\node [below = 5.25 of EV.center, anchor = center] {\footnotesize \bf \em Symmetry: $\mb x \mapsto -\mb x$};
\node [below = 5.65 of EV.center, anchor = center] {\footnotesize $\bb G = \{ \pm 1 \}$};
\node [right=6.05 of P.center, anchor=north] (DL) {\bf Dictionary Learning};
\node [below = 2.5 of DL.center, anchor = center] {\includegraphics[width=1.15in]{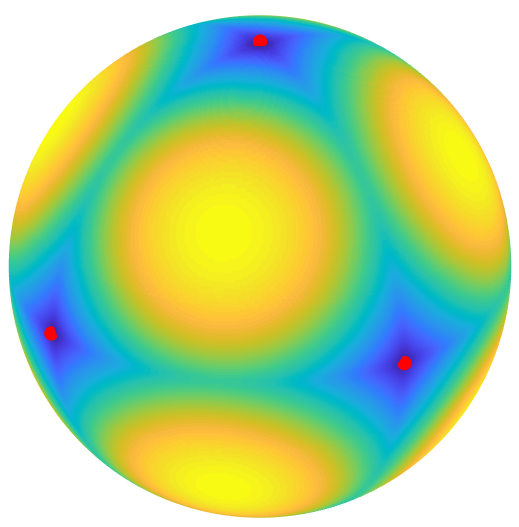}};
\node [below = .50 of DL.north, anchor = north] {\footnotesize \em Approximate a given matrix $\mb Y$ }; 
\node [below = .80 of DL.north, anchor = north] {\footnotesize \em as $\mb Y \approx \mb A \mb X$, with $\mb X$ sparse};
\node [below = 4.55 of DL.center, anchor = center] {\footnotesize   $\min_{\, \mb A \in \mathcal A,\mb X} \tfrac{1}{2} \| \mb Y - \mb A \mb X \|^2_F +  \lambda \| \mb X \|_1$.};
\node [below = 5.25 of DL.center, anchor = center] {\footnotesize \bf \em Symmetry: $(\mb A,\mb X) \mapsto (\mb A\mb \Gamma, \mb X \mb \Gamma^{*})$};
\node [below = 5.65 of DL.center, anchor = center] {\footnotesize $\bb G = \mr{SP}(n)$};
\node [below = 6 of EV.center, anchor=north] (TD) {\bf Tensor Decomposition};
\node [below = 2.5 of TD.center,anchor=center] {\includegraphics[width=1.15in]{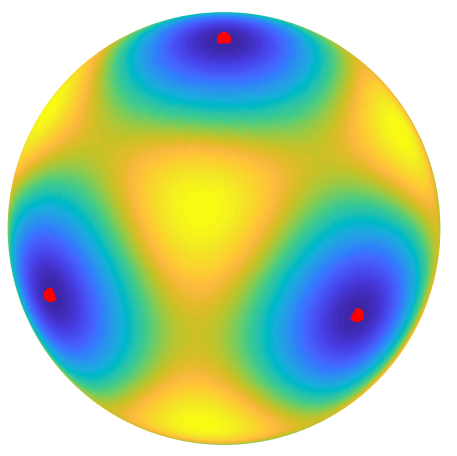}};
\node [below = .47 of TD.north, anchor = north] {\footnotesize \em Determine components $\mb a_i$ of an orthogonal}; 
\node [below = .80 of TD.north, anchor = north] {\footnotesize \em  decomposable tensor $\mb T = \sum_i \mb a_i \otimes \mb a_i \otimes \mb a_i \otimes \mb a_i$};
\node [below = 4.55 of TD.center, anchor = center] {\footnotesize   $\max_{\mb X \in O(n)} \sum_i \mb T(\mb x_i,\mb x_i,\mb x_i,\mb x_i)$.};
\node [below = 5.25 of TD.center, anchor = center] {\footnotesize \bf \em Symmetry: $\mb X \mapsto \mb X \mb \Gamma$};
\node [below = 5.65 of TD.center, anchor = center] {\footnotesize $\bb G = \mr{P}(n)$};
\node [below = 6 of DL.center, anchor= north] (SSD) {\bf Short-and-Sparse Deconvolution};
\node [below = 2.5 of SSD.center, anchor = center] {\includegraphics[width=1.15in]{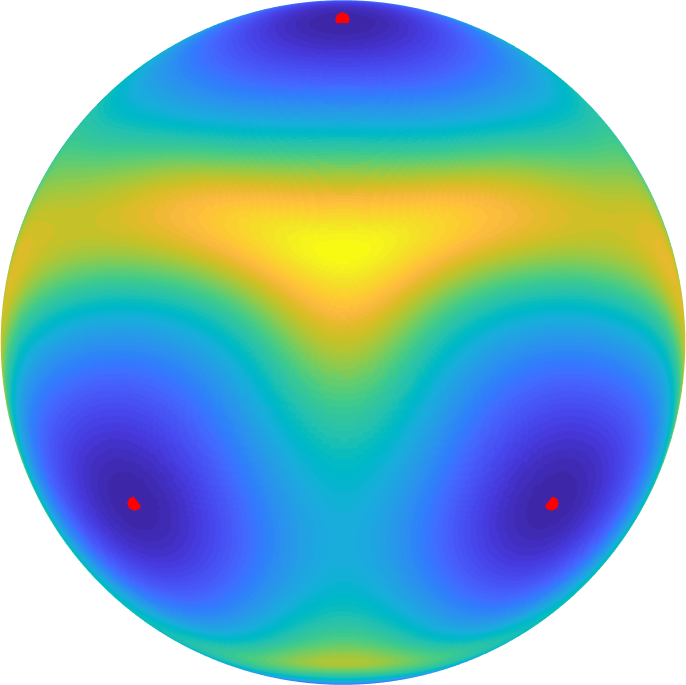}};
\node [below = .47 of SSD.north, anchor = north] {\footnotesize \em Recover a short $\mb a$ and a sparse $\mb x$}; 
\node [below = .80 of SSD.north, anchor = north] {\footnotesize \em from their convolution $\mb y = \mb a \ast \mb x$.};
\node [below = 4.55 of SSD.center, anchor = center] {\footnotesize   $\min_{\mb a, \mb x} \tfrac{1}{2} \|  \mb y -\mb  a \ast \mb x \|_2^2  \;+\; \lambda \| \mb x \|_1$.};
\node [below = 5.25 of SSD.center, anchor = center] {\footnotesize \bf \em Symmetry: $(\mb a,\mb x) \mapsto (\alpha s_{\tau}[ \mb a], \alpha^{-1} s_{-\tau}[\mb x])$};
\node [below = 5.65 of SSD.center, anchor = center] {\footnotesize $\bb G = \bb Z_n \times \bb R_*$ or $\bb G = \bb Z_n \times \set{\pm 1}$};
\end{tikzpicture}
}
\caption{Four examples of problems with discrete symmetries. We discuss this family of problems in more detail in Section \ref{sec:discrete}.} \label{fig:discrete-symmetries}
\end{figure}

In this paper, we identify two families of symmetric nonconvex problems, which exhibit similar geometric characteristics. The first family of problems exhibit {\em rotational symmetries}: the group $\bb G$ is either an orthogonal group $O(n)$ or $SO(n)$.\footnote{Here, $SO(n)$ is a subgroup of $O(n)$, with determinant equalling to unity.} The phase retrieval problem described above is a canonical example; {\bf Figure \ref{fig:rotational-symmetries}} illustrates this family. The second family of problems exhibit {\em discrete symmetries}: signed permutations $\mathrm{SP}(n)$, signed shifts $\bb Z_n \times \{ \pm 1 \}$, or products of these. The dictionary learning problem discussed above is a canonical example; {\bf Figure \ref{fig:discrete-symmetries}} shows several others.

In the remainder of this paper, we explore the geometry of these two families of problems in more depth. \Cref{sec:rotational} studies problems with rotational symmetries, starting with a very simple model problem in which the goal to recover a single complex scalar from magnitude measurements, and drawing conclusions that carry over to more complicated measurement models for phase retrieval \cite{candes2013phase-matrix,candes2015phase,shechtman2015phase,sun2018geometric,fannjiang2020numerics} and related problems in low-rank matrix factorization and recovery \cite{ge2016matrix,ge2017no,chi2019nonconvex}. \Cref{sec:discrete} studies problems with discrete symmetries, starting again from another simple model problem and extracting conclusions that carry over to problems such as dictionary learning \cite{sun2017complete1,sun2017complete2,gilboa2018efficient,qu2019analysis}, blind deconvolution \cite{ling2017blind,zhang2018structured,kuo2019dq,lau2019short,li2018global,qu2019blind} and tensor decomposition \cite{ge2015escaping,ge2017optimization}. As mentioned above, this area is rich with open problems; we highlight a few of these in \Cref{sec:discussion}. These open problems span high-dimensional geometry and algorithms. Nevertheless, our main focus throughout this survey is geometric: we will concentrate on the connections between symmetry and geometry. As described above, these geometric analyses have strong implications: in many cases, they guarantee that problems can be solved globally in polynomial time. For problems with rotational symmetry, we recommend the (complementary) survey papers \cite{chen2018harnessing,chi2018nonconvex,jin2021nonconvex} for a more detailed exposition of issues at the interface of statistics and computation; \Cref{sec:discussion} also briefly discusses similar considerations for problems exhibiting discrete symmetries, where we refer readers to our companion overview paper \cite{qu2020finding} for more computational and application aspects on these problems. On the other hand, in order to keep the development of this overview focused on geometric intuitions, we will only treat computational issues at a high level. For developments on 1st and 2nd order optimization methods related to this topic, we refer readers to recent works \cite{jain2017non,hu2020brief,manton2020geometry,jin2021nonconvex,danilova2020recent,boumal2022intromanifolds}. In particular, \cite{jain2017non} focuses on alternating direction type of methods and local analysis for nonconvex optimization problems in machine learning; the recent works \cite{danilova2020recent,hu2020brief,jin2021nonconvex,boumal2022intromanifolds} focus on reviews of efficient nonconvex optimization methods, including zero-th order \cite{danilova2020recent}, first-order \cite{jin2021nonconvex} and second-order \cite{danilova2020recent} methods, and Riemannian optimization methods \cite{hu2020brief,manton2020geometry,boumal2022intromanifolds}. As mentioned above, for many naturally occurring problems in signal processing and machine learning, these methods not only find critical points, they actually find global minimizers. The geometric considerations that we introduce below help explain why this is the case!

\paragraph{Basic notations.} Before proceeding, we recap and introduce some basic notations. Throughout the paper, all vectors/matrices are written in bold font $\mb a$/$\mb A$; indexed values are written as $a_i, A_{ij}$. For a matrix $\mb A$, we use $\mb A^*$ to denote the transpose of $\mb A$ (conjugate transpose if $\mb A$ is complex). We use $\bb S^{n-1}$ to denote an $n$-dimensional unit sphere in the Euclidean space $\bb R^n$. We let $[m] =\Brac{1,2,\cdots,m}$. We use $\odot$ to denote the entry-wise Hadamard product, and we use $\ast$ to denote linear convolution. 
For any given vector $\mb a$, we use $\norm{\mb a}{p} = \paren{ \sum_{i=1} \abs{a_i}^p}^{1/p}$ to denote its $\ell_p$-norm. For any given matrix $\mb A$, we use $\norm{\mb A}{F}$ and $\norm{\mb A}{}$ to denote its Frobenius norm and spectral norm, respectively.

\section{Nonconvex Problems with Rotational Symmetry}
\label{sec:rotational}

In this section, we study the first main class of problems in our taxonomy of symmetric nonconvex problems: problems with rotational symmetry. This class includes important model problems in phase recovery \cite{shechtman2015phase,fannjiang2020numerics} and low-rank estimation \cite{chi2019nonconvex}. We begin by developing a few basic intuitions through a toy phase retrieval problem; we then show how these intuitions help to explain the geometry of a range of problems from imaging and machine learning. 

\begin{figure}[t]
\centerline{
\begin{tikzpicture}
\node at (-.5,0) {\includegraphics[width=2.5in]{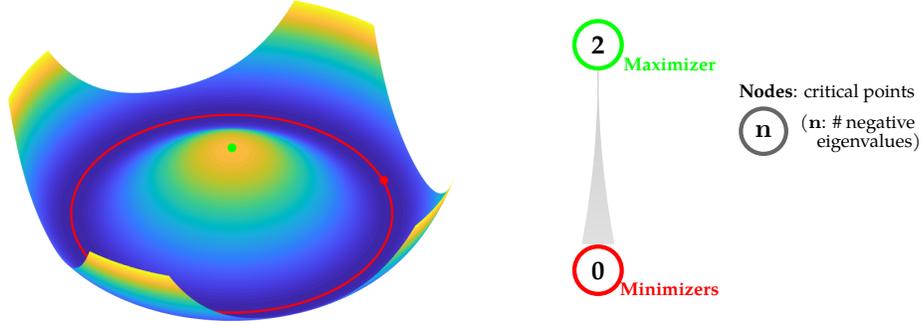}};
\node[shape=circle,draw = green, ultra thick] (max) at (4.5,1.5) {\bf 2};
\node[shape=circle,draw = red, ultra thick] (min) at (4.5,-1.5) {\bf 0};
\node at (5.45,1.25) {\color{green}\scriptsize \bf Maximizer};
\node at (5.45,-1.75) {\color{red}\scriptsize \bf Minimizers};
\node[anchor = north west] at (6.25,1.1) {\scriptsize {\bf Nodes}: critical points};
\node [shape=circle,draw = black!60!white, ultra thick] at (6.7,0.35) {$\mathbf n$};
\node at (7.95,0.45) {\scriptsize ($\mathbf{n}$: \# negative};
\node at (8.1,0.2) {\scriptsize eigenvalues)};
\draw[line width = .1pt, decoration={width and color change,start color=black!20!white,end color=black!60!white}, decorate]  (max.south) to (min.north); 
\end{tikzpicture}
}
\caption{{\bf Phase Retrieval with a Single Unknown.} We plot the objective function $\varphi(x)$ for phase retrieval with a single complex unknown. All {\color{red} local minimizers (red)} are symmetric copies $x_0 e^{\mf i \phi}$ of the ground truth $x_0 \in \bb C$. There is also a {\color{green}local maximizer (green)} at $x = 0$; at this point, $\varphi$ exhibits negative curvature in directions that break symmetry. Right: critical points arranged according to objective function $\varphi$, labelled according to their index (number of negative eigenvalues).} \label{fig:pr-1c}
\end{figure}

\subsection{One Minimal Example: Phase Retrieval with a Single Unknown}\label{sec:minimal-pr}

We first consider a model problem, in which our goal is to recover a single complex scalar $x_0 \in \bb C$ from $m$ magnitude measurements
\begin{align}
y_1 = \abs{a_1 x_0 }, \, \dots, \, 
y_m = \abs{a_m x_0}, 
\end{align}
where $a_1, \dots, a_m \in \bb C$ are known complex scalars. Collecting our observations $y_i$ into a single vector $\mb y \in \bb R^m$ and collecting the $a_i$ into a single vector $\mb a \in \bb C^m$, we can express this measurement model more compactly as 
\begin{equation} \label{eqn:gpr-simple}
\mb y \;=\; \abs{ \mb a x_0 }.
\end{equation}
Our goal is to determine $x_0$, up to a phase. This is a heavily simplified (indeed, trivialized!) version of the {\em generalized phase retrieval} problem \cite{candes2013phaselift,candes2015phase,sun2018geometric}, which we will describe in more detail in \Cref{sec:pr}. Here our goal is simply to understand the consequences of the phase symmetry of the measurement model \eqref{eqn:gpr-simple} for optimization. To this end, we study a model optimization problem,
\begin{align} \label{eqn:pr-1c}
\min_{x \in \bb R}\quad \varphi(x) \;\doteq\; \tfrac{1}{2} \bigl\|  \, \mb y^2-\abs{\mb a x}^2 \, \bigr\|_2^2,
\end{align}
which minimizes the sum of squared differences between the squared magnitudes of $\mb a x$ and those of $\mb a x_0$. Note that 
\begin{equation}
\varphi(x) = \tfrac{1}{4} \| \mb a \|_4^4 \paren{ |x|^2 - |x_0|^2 }^2.
\end{equation}
This is a function of a complex scalar $x = x_r + \mf i x_i$. We can study its geometry by identifying $x$ with a two-dimensional real vector $\bar{\mb x} = (x_r,x_i)$. The slope and curvature of the function $\varphi(\bar{\mb x})$ are captured by the gradient and Hessian, 
\begin{align}
\nabla \varphi(\bar{\mb x}) \;&=\; \| \mb a \|_4^4 \Bigl( | x |^2 - |x_0|^2 \Bigr) \left[ \begin{array}{c} x_r \\ x_i \end{array} \right], \\
\nabla^2 \varphi(\bar{\mb x}) \;&=\; \| \mb a \|_4^4  \Bigl( \left( |x|^2 - |x_0|^2 \right) \mb I + 2 \bar{\mb x} \bar{\mb x}^* \Bigr).
\end{align}
\Cref{fig:pr-1c} visualizes the objective $\varphi(\cdot)$ and its critical points. By setting $\nabla \varphi = 0$, and inspecting the Hessian, we obtain that there exist two families of critical points: global minimizers at $x = x_0 e^{\mf i \phi}$, and a global maximizer at $x = 0$. We notice that:

\begin{itemize}
\item {\bf \em Symmetric copies of the ground truth are minimizers}. The points $\Brac{x_0 e^{\mf i \phi}}$ are the only local minimizers. In problems with phase ambiguities, we expect a circle $O(2) \cong \bb S^1$ of minimizers. In addition, the Hessian is positive semidefinite, but rank deficient at the global minimizers: the zero curvature direction (along which the objective $\varphi$ is flat) is precisely the direction that is tangent to the set of equivalent solutions $g \cdot {\mb x}_\star$ at ${\mb x}_\star$. Normal to this set, the objective function exhibits positive curvature -- a form of restricted strong convexity.
\item {\bf \em Negative curvature in symmetry breaking directions}. There is a local maximizer at $x = 0$, which is equidistant from the target solutions $\set{ x_0 e^{\mf i \phi} }$. At this point $\nabla^2 \varphi \prec \mb 0$; there is negative curvature in every direction, and movement in any direction breaks the symmetry. 

\end{itemize}

\subsection{Generalized Phase Retrieval}\label{sec:pr}

The univariate phase retrieval problem is an extreme idealization of a basic problem in imaging: recovering a signal from phaseless measurements \cite{candes2013phase-matrix,shechtman2015phase}. This problem arises in many application areas, including electron microscopy \cite{jianwei2002high}, diffraction and array imaging \cite{bunk2007diffractive,anwei2011array}, acoustics \cite{balan2006signal,balan2010signal}, quantum mechanics \cite{Corbett2006pauli,reichenbach1965philosophic} and quantum information \cite{heinosaari2013quantum}, where the goal is to image complex molecular structures. Illuminating a sample with coherent light produces a diffraction pattern, which is approximately the Fourier transform of the sample's density. If we could measure this diffraction pattern, we could recover an image of the sample with atomic resolution, simply by inverting the Fourier transform. However, there is a wrinkle: typically, the magnitude of the Fourier transform is much easier to measure than the phase -- the magnitude can be measured by aggregating energy over time, whereas measuring the phase of a high frequency signal requires the detector to be sensitive to very rapid changes. The Fourier phase retrieval problem asks us to reconstruct a complex signal from magnitude measurements only:
\begin{align*}
   \text{find} \;\mb x \quad \text{such that} \quad \abs{\mc F[\mb x] } \;= \; \mb y.
\end{align*}
This problem is widespread in scientific imaging \cite{millane1990phase,robert1993phase, walther1963question, fienup1987phase}, but it is quite challenging: it is ill-posed in one dimension, and in higher dimensions even the most effective numerical methods remain sensitive to initialization and parameter tuning \cite{fienup2013phase}, where  for more details we refer readers to recent survey papers~\cite{shechtman2015phase,jaganathan2015phase,fannjiang2020numerics}. From the perspective of this survey, one explanation for this difficulty resides in the symmetries of the measurement operator $| \mc F[\cdot]|$: in addition to phase symmetry, the mapping $\mb x \mapsto |\mc F[\mb x] |$ is invariant under shifts and conjugate reversal of the signal $\mb x$. 

In recent years, the applied mathematics community has investigated variants of the above problem in which the Fourier transform $\mc F(\cdot)$ is replaced by a more general linear operator $\mc A(\cdot)$ \cite{candes2013phaselift,candes2013phase-matrix,candes2015diffraction}. A ``generic'' map $\mb x \mapsto |\mc A[\mb x]|$ has simpler symmetries -- typically only a phase symmetry, $| \mc A [ \mb x e^{j \phi} ] | = | \mc A [\mb x] |$. This makes generic phase recovery problems easier to study and easier to solve. While the Fourier model is more widely applicable to physical imaging, the generic phase retrieval model does capture aspects of certain less conventional imaging setups, including ptychography \cite{yeh2015experimental,jaganathan2016stft,pfeiffer2018x} (i.e., $\mc A(\cdot)$ is the Short Time Fourier Transform), coded illuminations \cite{tian20153d,kellman2019physics}, and coded diffraction patterns \cite{candes2015phase}. A model one-dimensional\footnote{Here, in comparison to the univariate case, one-dimensional means $\mb x \in \bb C^n$ is a vector instead of a matrix.} version of the generalized phase retrieval problem can be formulated as follows:
\begin{equation}
   \text{find} \quad \mb x \in \bb C^n\quad \text{such that} \quad \left| \mb A \mb x \right| = \mb y,
\end{equation}
where $\mb A \in \bb C^{m \times n}$ is a matrix which represents the sensing process.

Analogous to the univariate phase retrieval in \eqref{eqn:pr-1c}, we can attempt to recover $\mb x_0$ by minimizing its misfit to the observed data $\mb y$, by solving 
\begin{align}\label{eqn:pr-4th}
   \min_{\mb x \in \bb C^n} \varphi(\mb x) \; \equiv\; \frac{1}{4m} \sum_{k=1}^m  \paren{ y_k^2 - \abs{ \mb a_k^* \mb x }^2 }^2,
\end{align}
where $\mb a_1, \dots, \mb a_m \in \bb C^n$ are the row vectors of $\mb A$. We saw from above that the univariate version of this function has a very simple landscape, which is dictated almost entirely by phase symmetry, and that it has no spurious local minimizers. {\em Should we expect similar behavior in this higher dimensional setting?}

\begin{figure}[t]
\centerline{
\begin{tikzpicture}
\node at (-5,0) {\includegraphics[width=1.75in]{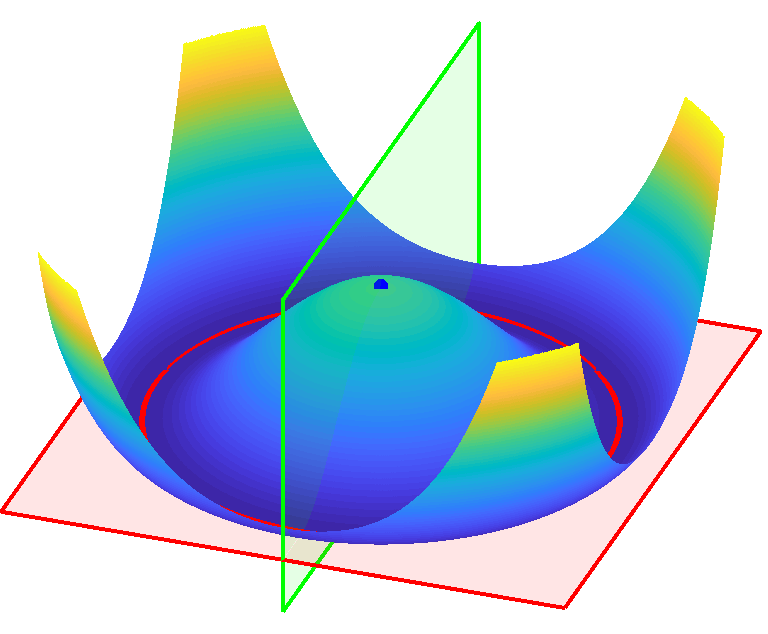}};
\node at (0,0) {\includegraphics[width=1.75in]{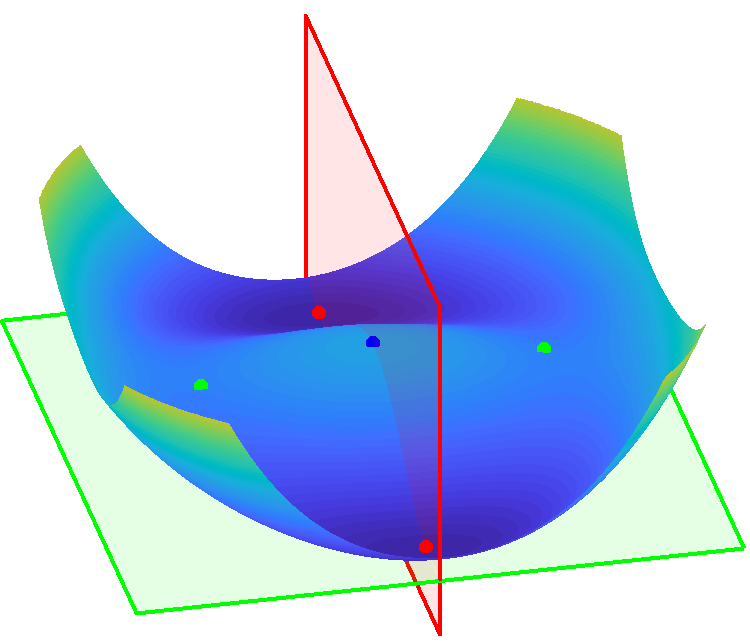}};
\node[shape=circle,draw = blue, ultra thick] (max) at (3.5,1.5) {$\mathbf{2n}$};
\node[shape=circle,draw = green, ultra thick] (saddle) at (4.5,.5) {$\mathbf{1}$};
\node[shape=circle,draw = red, ultra thick] (min) at (3.5,-1.5) {$\mathbf{0}$};
\node at (4.55,1.25) {\color{blue}\scriptsize \bf Maximizer};
\node at (5.25,.25) {\color{green}\scriptsize \bf Saddles};
\node at (4.45,-1.75) {\color{red}\scriptsize \bf Minimizers};
\node[anchor = north west] at (6,1.1) {\scriptsize {\bf Nodes}: critical points};
\node [shape=circle,draw = black!60!white, ultra thick] at (6.45,0.35) {$\mathbf n$};
\node at (7.7,0.45) {\scriptsize ($\mathbf{n}$: \# negative};
\node at (7.85,0.2) {\scriptsize eigenvalues)};
\node[anchor = north west] at (6,-.4) {\scriptsize {\bf Connections}};
\node[anchor = west] at (6.45,-.995) {\scriptsize {Dispersive}};
\draw[line width = .25pt, decoration={width and color change legend,start color=black!60!white,end color=black!60!white}, decorate] (6.4,-.85) to (6.4,-1.20);
\draw[line width = .1pt, decoration={width and color change,start color=black!20!white,end color=black!60!white}, decorate]  (max.south) to (min.north); 
\draw[line width = .1pt, decoration={width and color change,start color=black!20!white,end color=black!60!white}, decorate] (max.south) to (min.north); 
\draw[line width = .1pt, decoration={width and color change,start color=black!20!white,end color=black!60!white}, decorate] (max) to (saddle);
\draw[line width = .1pt, decoration={width and color change,start color=black!20!white,end color=black!60!white}, decorate] (saddle) to (min);
\end{tikzpicture}
}
\caption{{\bf Generalized Phase Retrieval.} We plot two slices of the landscape of the generalized phase retrieval problem with Gaussian measurements. Left: slice containing symmetric copies of the ground truth $\mb x_0 e^{\mathfrak{i} \phi}$. Middle slice containing {\color{red} minimizers} $\mb x_0$, $-\mb x_0$ and one orthogonal direction. Notice that at both the {\color{blue} maximizer} and {\color{green} saddle points}, there is negative curvature in the direction that breaks symmetry between $\mb x_0$ and $-\mb x_0$. Right: critical points arranged according to objective $\bb E \varphi$, labeled with their indices (number of negative eigenvalues). Connections between critical points are ``dispersive'': downstream negative curvature directions are the image of upstream negative curvature directions under gradient flow (see Appendix \ref{app:dispersive}).} \label{fig:gpr}
\end{figure}
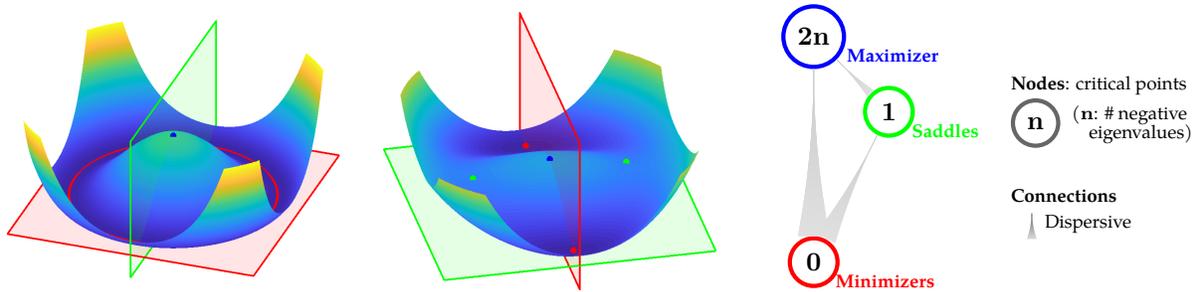

\subsection*{Geometry of Generalized Phase Retrieval}

One way of generating intuition is to assume that the sampling vectors $\Brac{\mb a_i}_{i=1}^m$ are chosen at random, and analyze $\varphi(\mb x)$ using tools from statistics. \Cref{fig:gpr} visualizes $\varphi(\mb x)$ when the $\mb a_k$ are Gaussian vectors\footnote{Formally, $\mb a_k$ are independent random vectors, with $\mb a_k = \mb a_k^r + \mf i \mb a_k^i$ with $\mb a_k^r$ and $\mb a_k^i$ independent iid $\mc N(0,\tfrac{1}{2})$.} and $m$ is large. As $m \to \infty$, $\varphi(\mb x)$ converges to its expectation $\bb E_{\mb a}\brac{ \varphi}$, which can be calculated in closed form. In \Cref{fig:gpr} (left), we can see the characteristic phase symmetry, identical to our univariate example above. However, this problem is higher dimensional. \Cref{fig:gpr} (center) plots the objective over a two-dimensional slice containing the ground truth and an orthogonal direction, where we observe:
\begin{itemize}
 	\item {\bf \em Symmetric copies of the ground truth are minimizers}. All the local minimizers are on the circle of points $\Brac{\mb x_0 e^{\mf i \phi}}$, which corresponds to the ground truth up to the (rotational) phase symmetry. Problems with higher dimensional symmetries have larger sets of minimizers -- e.g., $O(r)$ symmetry leads to a manifold of minimizers that is isometric to $O(r)$. 
	\item {\bf \em Negative curvatures in symmetry-breaking directions}. In higher dimensional examples, we encounter a variety of local maximizers, saddle points, etc. Nevertheless, these critical points occur near balanced superpositions of equivalent solutions, and exhibit a negative curvature in directions $\pm \mb x_0$, which breaks the symmetry.
	\item {\bf\em Cascade of saddle points}. As shown schematically in \Cref{fig:gpr}, the critical points can be graded based on the number of negative eigenvalues of the Hessian: critical points with higher objective values have more negative eigenvalues. Moreover, the objective has a ``dispersive'' property: upstream negative curvature discourages stagnation near the stable manifold of downstream critical points (see Appendix \ref{app:dispersive} for more detail). 
\end{itemize}

\subsection*{Finite Samples, Structured Measurements, Different Objectives}

The exposition in the previous section is still quite idealized: the measurements are Gaussian, and we have infinitely many of them. Moreover, we have assumed a particular objective $\varphi(\mb x)$, which is not the common objective used in practice. Fortunately, the qualitative conclusions of the previous subsection carry over to more structured and challenging settings for generalized phase retrieval.\footnote{But {\em not} to the Fourier model, which has different symmetries. We discuss challenges and open problems around Fourier measurements in \Cref{sec:discrete} and \Cref{sec:discussion}.} In the following, we briefly describe these extensions, while noting technical caveats and open problems. 

\paragraph{Finite Samples} Phase retrieval is a sensing problem; measurements cost resources. It is important to minimize the number of measurements $m$ required to accurately reconstruct $\mb x$. Under the Gaussian model, the particular loss function $\varphi(\cdot)$ in \eqref{eqn:pr-4th} is a sum of independent heavy-tailed random variables. Relatively straightforward considerations show that when $m \gtrsim n^2$, gradients and hessians concentrate uniformly about their expectations, and the objective has no spurious local minimizers. This number of samples is clearly suboptimal -- $n^2$ measurements to recover about $n$ complex numbers. The challenge is that the objective function \eqref{eqn:pr-4th} contains fourth moments of Gaussian variables, and its distribution is therefore heavy-tailed. Using arguments that are tailored to this situation, the required number of samples can be improved to $m \gtrsim n \log^3 n$ \cite{sun2018geometric}. Moreover, modifying the objective \eqref{eqn:pr-4th} to remove large terms (ala robust statistics) can improve this to essentially optimal ($m \gtrsim n$) \cite{chen2017solving}.\footnote{Other approaches to producing analyses with small sample complexity include restricting the analysis to a small neighborhood of the ground truth, and initializing in this neighborhood using spectral methods that leverage the statistics of the measurement model \cite{candes2015phase,waldspurger2015phase}, or forgoing uniform geometric analysis and directly reasoning about trajectories of randomly initialized gradient descent \cite{ma2017implicit}.}

\paragraph{Different Objective Functions} The ``squares of the squares'' formulation in \eqref{eqn:pr-4th} is smooth and hence simple to analyze, but is typically not preferred in practice, especially when observations are noisy. Alternatives include 
$\varphi(\mb x) = \sum_i \left| y_i^2 - |\mb a_i^* \mb x|^2 \right|$ \cite{wang2017solving,duchi2019solving}, $\varphi(\mb x) = \sum_i \left| y_i - |\mb a_i^* \mb x| \right|^2$ \cite{davis2017nonsmooth}, and maximum likelihood formulations that model (Poisson) noise in the observations $y_i$  \cite{chen2017solving,li2021algorithms}. Although these formulations differ in details, the major features of the objective landscape are independent of the choice of $\varphi$. For Gaussian sensing vectors $\Brac{\mb a_i}_{i=1}^m$, the expectation $\bb E_{\mb a} \Brac{\varphi}$ has no spurious minimizers; moreover, all objectives have a minimizer at zero and a family of saddle points orthogonal to $\mb x_0$. However, proving (or disproving) that these objectives have benign global geometry for finite $m$ is still an open problem. Existing small sample analyses \cite{chen2017solving,wang2017solving,davis2017nonsmooth} control the behavior of the objective in a neighborhood of $\mb x_0 e^{\mf i \phi}$, and initialize in this neighborhood using statistical properties of the measurement model. 

\paragraph{Structured Measurements} Geometric intuitions for Gaussian $\mb A$ carry over to several models that are more closely connected with imaging practice. Examples include convolutional models, in which we observe the modulus of the convolution $\mb y = | \mb a \ast \mb x |$ of the unknown signal $\mb x$ with a known sequence $\mb a$ \cite{qu2017convolutional} and coded diffraction patterns, in which we make multiple observations $\mb y_l = | \mc F [ \mb d_l \circ \mb x ] |$, where $\circ$ denotes an elementwise product \cite{candes2015diffraction}. If the filter $\mb a$ or the masks $\mb d_l$ are chosen at random from appropriate distributions, these structured measurements yield the same asymptotic objective function $\bb E \varphi$. In particular, in the large sample limit (i.e., infinite long filter $\mb a$ for the convolutional model, or infinitely many diffraction patterns $\mb d_l$ in the coded diffraction model), these measurements still lead to optimization problems with no spurious local minimizers. Similar to the situation with nonsmooth objective functions, the best known theoretical sample complexities are obtained by initializing near the ground truth, using statistical properties of $\mb A$. Globally analyzing structured measurements in the small sample regime is a challenging open problem.

\vspace{.1in}

\noindent The above discussion only scratches the surface of the growing literature on generalized phase retrieval, we refer readers to \cite{shechtman2015phase,jaganathan2015phase,fannjiang2020numerics,grohs2020phase} for a more comprehensive survey on recent developments. From the perspective of this survey, the unifying thread through all of these models, objectives and problems is the simple model geometry in \Cref{fig:gpr}. In the next section, we will see a similar phenomenon with low-rank matrices: a model geometry originating in matrix factorization recurs across a sequence of increasingly challenging matrix recovery problems.


\subsection{Low Rank Matrix Recovery}
\label{sec:lowrank}

The problem of recovering a low-rank matrix from incomplete and unreliable observations finds broad applications in robust statistics, recommender systems, data compression, computer vision, and so on \cite{davenport2016overview}. In matrix recovery problems, the goal is to estimate a matrix $\mb X_0 \in \bb R^{n_1 \times n_2}$ from incomplete or noisy observations. Typically, this problem is ill-posed without any assumptions on the matrix $\mb X_0$. In many applications where the data are highly structured, $\mb X_0$ can be assumed to be {\em low rank}, or approximately so:
\begin{equation}
r = \mr{rank}(\mb X_0) \ll \min\left\{ n_1, n_2 \right\}.
\end{equation}
Additionally, any rank-$r$ matrix can be expressed as a product of a tall $n_1 \times r$ matrix and a wide $r \times n_2$ matrix:
\begin{align} \label{eqn:fac}
\mb X_0 =\mb U\mb V^*,\quad\mb U\in\R^{n_1\times r},\mb V\in\R^{n_2\times r}.
\end{align}
A very popular strategy for recovering $\mb X_0$ is to start with some objective function $\psi(\mb X)$ that enforces consistency with observed data, and then parameterize $\mb X$ in terms of the factors $\mb U$, $\mb V$ \cite{burer2003nonlinear}, yielding the optimization problem 
\begin{equation} \label{eqn:UVnoreg}
\min_{\mb U,\mb V} \, \varphi(\mb U,\mb V) \equiv \psi(\mb U\mb V^*),
\end{equation}
where we shall discuss the concrete form of the loss $\varphi(\cdot)$ later in this section (i.e., see \eqref{eqn:symmetric-factorization}).

\subsection*{Symmetries of Low Rank Models} 
Formulations like \eqref{eqn:UVnoreg} are almost always nonconvex in the factors $\mb U$ and $\mb V$, due to symmetries of the factorization \eqref{eqn:fac}. Indeed, for any invertible matrix $\mb \Gamma \in \bb R^{r \times r}$, 
\begin{equation}
\mb U \mb V^* \,=\, \mb U \mb \Gamma \mb \Gamma^{-1} \mb V^* \,=\, \left( \mb U \mb \Gamma \right) \left( \mb V \mb \Gamma^{-*} \right)^*
\end{equation}
Because of this ambiguity, the problem \eqref{eqn:UVnoreg} always possess a {\em general linear} (invertible matrix) symmetry:
\begin{equation}
(\mb U, \mb V) \equiv (\mb U\mb \Gamma, \mb V \mb \Gamma^{-*}), \qquad \mb \Gamma \in \mr{GL}(r). \label{eqn:GL}
\end{equation}
As the determinant of a general linear matrix $\mb \Gamma$ can be arbitrarily close to zero, and hence be arbitrarily ill-conditioned, the equivalence class of solutions $(\mb U,\mb V)$ has somewhat complicated geometry, as a subset of $\bb R^{n_1 \times r} \times \bb R^{n_2 \times r}$.\footnote{For example, it is neither closed nor bounded.} Fortunately, as we shall see in the following, it is not difficult to reduce this general linear symmetry to a simpler and better conditioned orthogonal symmetry, either by using information about the target $\mb X_0$, or by adding additional regularization terms to \eqref{eqn:UVnoreg}. 

\paragraph{Rotational Symmetries for Symmetric $\mb X_0$} If we have extra information that the target solution $\mb X_0$ is {\em symmetric and positive semidefinite}, then it admits a factorization of the form $\mb X_0 = \mb U_0 \mb U_0^*$. Thus, we can take $\mb U = \mb V$, which gives a slightly simpler problem 
\begin{equation} \label{eqn:UUnoreg}
\min_{\mb U} \varphi(\mb U) \equiv \psi(\mb U \mb U^*),
\end{equation}
with a smaller symmetry group. For any $\mb \Gamma \in O(r)$, 
$\mb U \mb U^* = \mb U \mb \Gamma \mb \Gamma^* \mb U^* = \left(\mb U\mb \Gamma \right) \left( \mb U \mb \Gamma \right)^*$,
and so the symmetric problem \eqref{eqn:UUnoreg} exhibits an orthogonal/rotational symmetry that
$\mb U \equiv \mb U \mb \Gamma$, for $\mb \Gamma \in O(r)$.

\paragraph{Rotational Symmetries for General $\mb X_0$ via Penalization} For general (non-symmetric) matrices $\mb X$, it is possible to add additional regularizations to \eqref{eqn:UVnoreg} in such a way that the general linear symmetry reduces to an orthogonal symmetry. At a high level, the idea is to add a penalty $\rho(\mb U,\mb V)$ that enforces $\mb U^* \mb U \approx \mb V^* \mb V$; this prevents $\mb U$ and $\mb V$ from having vastly imbalanced scales.\footnote{For example, $\rho(\mb U, \mb V) = \tfrac{1}{2} \| \mb U^* \mb U - \mb V^* \mb V \|_F$ accomplishes this.} The penalty $\rho$ can be chosen such to be $O(r)$-symmetric, such that the combined problem
\begin{equation}
\min_{\mb U,\mb V} \; \varphi(\mb U,\mb V) + \rho(\mb U,\mb V),
\end{equation}
possesses a simple $O(r)$ symmetry: $(\mb U,\mb V) \equiv (\mb U \mb \Gamma,\mb V\mb \Gamma)$, for $\mb \Gamma \in O(r)$.

\paragraph{Model Problems and the Matrix Recovery Zoo.} There are many variants of the vanilla matrix recovery problem, which are motivated by different applications and impose different assumptions on the observations and the noise \cite{davenport2016overview,ge2017no,chi2019nonconvex}. Although these problems have their own technical challenges, they have certain qualitative features in common. At a high level, ``matrix {\em recovery} problems act like matrix {\em factorization} problems'' \cite{ge2017no}. In the next section, we will begin by describing in detail the geometry of matrix factorization, and then describe how these intuitions carry over to matrix recovery from incomplete or unreliable observations.

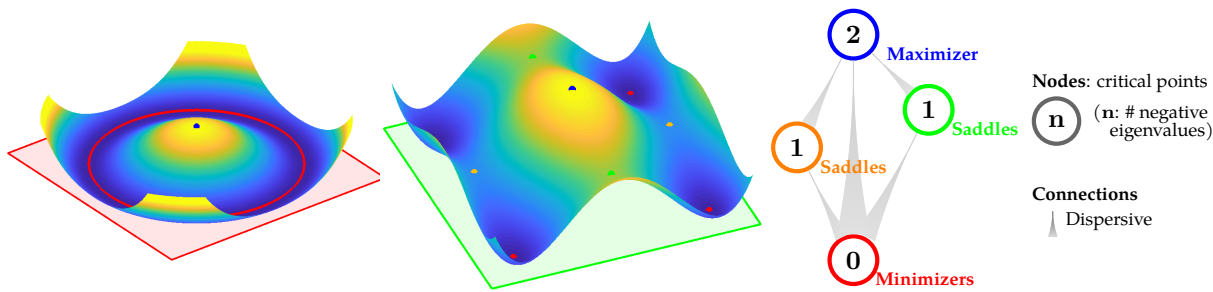
\begin{figure}[t]
\centerline{
\begin{tikzpicture}
\node at (-5,0) {\includegraphics[width=2.0in]{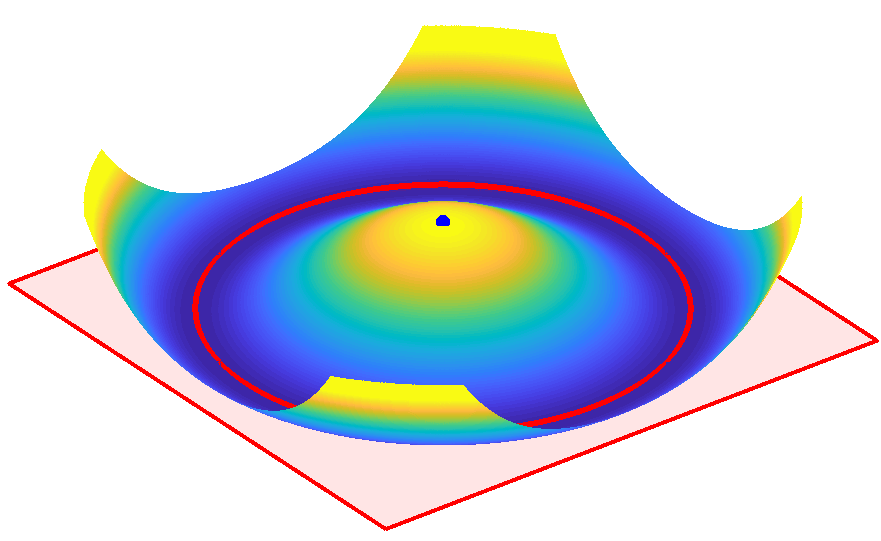}};
\node at ( 0,0) {\includegraphics[width=2.0in]{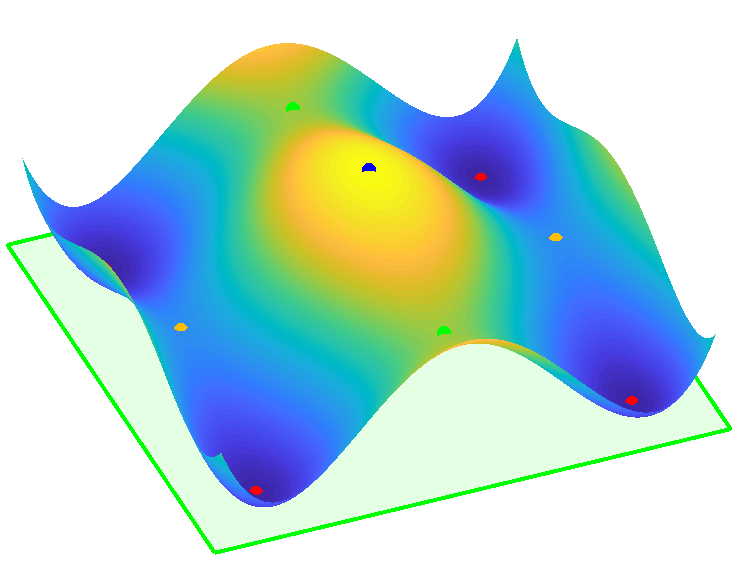}};
\node[shape=circle,draw = blue, ultra thick] (max) at (3.75,1.5) {$\mathbf{2}$};
\node[shape=circle,draw = green, ultra thick] (saddle) at (4.75,.5) {$\mathbf{1}$};
\node[shape=circle,draw = orange, ultra thick] (saddle2) at (3,0) {$\mathbf{1}$};
\node[shape=circle,draw = red, ultra thick] (min) at (3.75,-1.5) {$\mathbf{0}$};
\node at (4.8,1.25) {\color{blue}\scriptsize \bf Maximizer};
\node at (5.5,.25) {\color{green}\scriptsize \bf Saddles};
\node at (4.7,-1.75) {\color{red}\scriptsize \bf Minimizers};
\node[anchor = north west] at (6,1.1) {\scriptsize {\bf Nodes}: critical points};
\node [shape=circle,draw = black!60!white, ultra thick] at (6.45,0.35) {$\mathbf n$};
\node at (7.7,0.45) {\scriptsize ($\mathbf{n}$: \# negative};
\node at (7.85,0.2) {\scriptsize eigenvalues)};
\node[anchor = north west] at (6,-.4) {\scriptsize {\bf Connections}};
\node[anchor = west] at (6.45,-.985) {\scriptsize {Dispersive}};
\draw[line width = .25pt, decoration={width and color change legend,start color=black!60!white,end color=black!60!white}, decorate] (6.40,-.85) to (6.40,-1.20);
\draw[line width = .1pt, decoration={width and color change,start color=black!20!white,end color=black!60!white}, decorate] (max) to (saddle);
\draw[line width = .1pt, decoration={width and color change,start color=black!20!white,end color=black!60!white}, decorate] (max) to (saddle2);
\draw[line width = .1pt, decoration={width and color change,start color=black!20!white,end color=black!60!white}, decorate] (saddle2) to (min);
\draw[line width = .1pt, decoration={width and color change,start color=black!20!white,end color=black!60!white}, decorate] (saddle) to (min); 
\draw[line width = .1pt, decoration={width and color change,start color=black!20!white,end color=black!60!white}, decorate] (max.south) to (min.north); 
\node at (3.75,-.25) {\color{orange} \scriptsize \bf Saddles};
\end{tikzpicture}
}
\caption{{\bf Geometry of Matrix Factorization.} Geometry of a model problem in which the target $\mb X_0$ is a symmetric matrix of rank two, with eigenvalues $\tfrac{3}{4}$ and $\tfrac{1}{2}$. Left: plot of the objective $\varphi$ over a slice of the domain containing all optimal solutions. Center: two families of saddle points, corresponding to rank-one approximations. Right: objective value $\varphi$ versus index for the four families of critical points in this problem. Again the critical points are {\em graded}, in the sense that $\varphi$ decreases with decreasing index, and the paths between them are dispersive, in the sense that downstream negative curvature directions are the image of upstream negative curvature directions under gradient flow.} \label{fig:mf}
\end{figure}

\subsection*{Geometry of Matrix Factorization} 

Our first model problem starts with a complete, noise-free observation $\mb Y = \mb X_0$ of a symmetric, positive semidefinite matrix $\mb X_0 \in \reals^{n \times n}$ of rank $r < n$, and attempts to factor it as $\mb X_0 = \mb U \mb U^*$ by minimizing the misfit to the observed data \cite{li2019symmetry}:
\begin{align} \label{eqn:symmetric-factorization}
   \min_{\mb U \in \bb R^{n \times r}} \;\varphi(\mb U,\mb U)\;\doteq\; 	\tfrac{1}{4} \norm{ \mb Y - \mb U \mb U^* }{F}^2.
\end{align}
This is a nonconvex optimization problem, with orthogonal symmetry $\mb U \equiv \mb U \mb \Gamma$. \Cref{fig:mf} visualizes the objective landscape for this problem. It turns out that the critical points of $\varphi$ are dictated by the eigen-decomposition of the symmetric matrix $\mb X_0$ -- {\em every critical point $\mb U$ is generated by selecting and appropriately scaling a subset of the eigenvectors of $\mb X_0$, and then applying a right rotation $\mb U \mapsto \mb U\mb R$.} At a slogan level, critical points correspond to ``under-factorizations'' of the ground truth. By inspecting the hessian, we find that:
\begin{itemize} 
 	\item {\bf \em Symmetric copies of the ground truth are minimizers}. Local minimizers are the critical points which select all of the top $r$ eigenvectors, which correspond to the ground truth up to a rotation symmetry;
	\item {\bf \em Negative curvature in symmetry-breaking directions}. At a saddle point, there are strict negative curvatures in any directions which increase the number of top eigenvectors that participate. 
	\item {\bf \em Cascade of saddle points}. Saddle points are critical points selecting subsets of the top $r$ eigenvectors. These saddle points can be graded based on the number of selected eigenvectors. \footnote{A natural descent algorithm only visit at most $r$ saddle points whose trajectory depends on the containment of the active eigenvectors at those saddle points.}
\end{itemize}
\Cref{fig:mf} (center) visualizes these effects. This model geometry carries over to non-symmetric matrices. For example, considering a penalized low-rank estimation problem 
\begin{equation} 
 \min_{\mb U \in \bb R^{n_1 \times r}, \mb V \in \bb R^{n_2 \times r}} \; \varphi(\mb U, \mb V) \; \doteq\; \tfrac{1}{4} \norm{\mb Y - \mb U \mb V^* }{F}^2 + \rho(\mb U, \mb V),
 \end{equation}
 we obtain a problem with $O(r)$ symmetry. Critical points are generated by appropriately scaling subsets of the {\em singular} vectors of $\mb Y$ -- see Appendix \ref{app:factorization} for details on both of these geometries.

\subsection*{From Matrix Factorization to Matrix Recovery and Completion}

We next describe how precise geometric analyses of matrix factorization extend to a more realistic problem of recovering a low-rank matrix from incomplete and unreliable observations. As we shall see, matrix recovery problems often retain important qualitative features of matrix factorization. We will illustrate this phenomenon through several instances of a model recovery problem, in which we observe $m$ linear functions of an unknown matrix $\mb X_0 \in \reals^{n_1 \times n_2}$:
\begin{align}\label{eqn:matrix-sensing-model}
  y_i \;=\; \innerprod{\mb A_i}{\mb X_0}, \quad 1\leq i \leq m, 
\end{align}
and the goal is to recover $\mb X_0$. This model is flexible enough to represent matrix completion from missing entries \cite{candes2009exact}, as well as more exotic sensing problems \cite{recht2010guaranteed,davenport2016overview}. We can write this observation model more compactly by defining a linear operator $\mc A: \bb R^{n_1 \times n_2} \to \bb R^m$ with $\mc A(\mb X) := \brac{ \innerprod{\mb A_i}{\mb X} }_{1\leq i \leq m}$. In this notation, 
 \begin{equation}
 \mb y = \mc A(\mb X).
 \end{equation}
  If $m < n_1 n_2$, the number of observations is smaller than the number of unknowns, and the recovery problem is ill-posed. Fortunately, matrices encountered in applications have low-complexity structures; for instance, they are usually low-rank or approximately so. As above, a rank-$r$ matrix $\mb X_0$ admits a factorization $\mb X_0 = \mb U_0 \mb V_0^*$, that we can enforce this low-rank structure by directly recovering the factors $\mb U$ and $\mb V$, up to a rotational symmetry. A natural approach is to minimize the misfit to the observed data:
 \begin{align}\label{eqn:low-rank-obj1}
   \min_{\mb U, \mb V} \;\varphi(\mb U,\mb V)\;\doteq\; 	\frac{1}{4m} \sum_{i=1}^m \paren{ y_i - \innerprod{ \mb A_i }{ \mb U \mb V^* } }^2 + \rho(\mb U, \mb V) \,=\, \frac{1}{4m} \norm{ \mb y - \mc A(\mb U\mb V^*) }{2}^2 \;+\; \rho(\mb U, \mb V),
\end{align}
where as above $\rho$ is a regularizer that encourages the factors to be balanced.


\paragraph{Matrix Sensing} If $\mc A = \mc I$ is an identity operator, then \eqref{eqn:low-rank-obj1}  simply reduces to the factorization problem. In this special situation, the measurement operator $\mc A$ {\em exactly} preserves the geometry of {\em all} $n_1 \times n_2$ matrices, in the sense that $\| \mc A [ \mb X ] \|_2 = \| \mb X \|_F$ for all $\mb X$. However, when $\mc A$ is a generic operator rather than an identity map, and the number of measurements is small ($m < n_1 n_2$), the exact geometric preservation becomes impossible. Fortunately, as long as $\mc A$ {\em approximately} preserves the geometry of the {\em low-rank} matrices -- a much lower dimensional set \cite{park2016non,bhojanapalli2016global,zhu2018global,li2018non,li2019symmetry},  then \eqref{eqn:low-rank-obj1} still ``behaves like the factorization problem'', and hence can be used to recover $\mb X_0$.\footnote{This intuition can be formalized through the {\em rank restricted isometry property} (rank RIP) \cite{recht2010guaranteed,davenport2016overview}.} When this approximation is sufficiently accurate, there is a bijection between the critical points of the sensing problem \eqref{eqn:low-rank-obj1} and those of factorization, which preserves the index (number of negative eigenvalues). Under this condition, every local minimum of the sensing problem is global \cite{bhojanapalli2016global}.

\paragraph{Matrix Completion} The most practical and important instance of the general sensing model \eqref{eqn:low-rank-obj1} is the {\em matrix completion} problem \cite{candes2009exact}, in which the goal is to recover a low-rank matrix from a subset of its entries. This model problem arises, e.g., in collaborative filtering \cite{rennie2005fast,koren2009bellkor}, where the goal is to predict users' preferences for various products based a few observed preferences. Variants of this problem also appear in sensor networks (determining positions of sensors from a few distance measurements) \cite{biswas2006semidefinite,so2007theory}, imaging (recovering shape from illumination) \cite{wu2010robust,zhou2014low} and the geosciences \cite{yang2013seismic,kumar2015efficient}, just to name a few. 

Matrix completion also inherits the geometry of matrix factorization, with several technical caveats, which are consequences of the fact that it is challenging to recover $\mb X_0$ whose energy is concentrated on a small number of entries: if we fail to sample these important entries, we will fail to recover $\mb X_0$. This basic issue affects both for the well-posedness of the matrix completion problem and for our ability to solve it globally using nonconvex optimization. Local optimization methods could potentially become trapped in the region of the space in which $\mb U\mb V^*$ is nearly sparse, since the measurements do not effectively sense such matrices. One simple fix is to add an additional regularizer on the rows $\mb u_i$ and $\mb v_i$ of the factors, which encourages them to have small norm. This forces the energy of $\mb U \mb V^*$ to be spread across many entries.\footnote{In details, one can add a penalty $\rho_{\mr{mc}}(\mb U,\mb V) \;=\; \lambda_1 \sum_{i=1}^{n_1} \paren{ \norm{ \mb e_i^*\mb U }{2} - \alpha_1 }_+^4 + \lambda_2 \sum_{j=1}^{n_2} \paren{ \norm{ \mb e_j^*\mb V }{2} - \alpha_2  }_+^4$ to \eqref{eqn:low-rank-obj1}} In particular, Ge et al. \cite{ge2016matrix} proved that the resulting problem has a benign global geometry whenever we observe a sufficiently large random subset $\Omega$ and the target matrix $\mb X_0$ is itself not too concentrated on a few entries, in a precise technical sense.\footnote{Formally, $\mb X_0$ is $\mu$-incoherent, in the sense that for its compact SVD $\mb X_0 = \mb B\mb \Sigma\mb C^*$, $\| \mb e_i^* \mb B \|_2 \le \sqrt{\mu r / n_1}$ and $\| \mb e_j^* \mb C \|_2 \le \sqrt{\mu r / n_2 }$.}

\paragraph{Robust Matrix Recovery} Many data analysis problems confront the analyst with data sets that are not only incomplete, but also corrupted. Robust matrix recovery is the problem of estimating a low-rank matrix $\mb X_0$ from such an unreliable observation. Different models of corruption may be applicable in different application scenarios. For example, in imaging and vision, individual features (entries of the matrix) may be corrupted, e.g., due to occlusion \cite{candes2011robust,peng2012rasl}. This can be modeled as a sparse error: $\mb Y = \mb X_0 + \mb S_0$, with both $\mb X_0 = \mb U_0 \mb V_0^*$ {\em and } $\mb S_0$ unknown. We start from a natural formulation 
\begin{equation} \label{eqn:rpca}
\min_{\mb U, \mb V, \mb S} \tfrac{1}{2} \| \mb U \mb V^* + \mb S - \mb Y \|_F^2 +  g_s( \mb S ) + \rho_r( \mb U , \mb V),
\end{equation} 
where $g_s(\mb S)$ is a regularizer that encourages $\mb S$ to be sparse. Partially minimizing with respect to $\mb S$, we obtain 
\begin{equation}
\min_{\mb U,\mb V} \; \psi( \mb U\mb V^* - \mb Y ) + \rho_r(\mb U, \mb V),
\end{equation}
where $\psi(\cdot)$ is a new function that measures data fidelity -- e..g, if $g_s$ is the $\ell^1$ norm, then $\psi$ is a {\em Huber function} \cite{huber1992robust}. This is again a matrix factorization problem, but with a different loss $\psi( \mb U \mb V^* - \mb Y)$. While there are a number of open issues around the global (and even local! \cite{li2020nonconvex,charisopoulos2019low}) geometry of this problem, known results again suggest that for certain choices of $g_s$ and $\rho_r$ it indeed inherits the geometry of factorization \cite{chi2019nonconvex}. Similar to matrix completion, technical issues arise due to the possibility of encountering low-rank matrices $\mb U \mb V^*$ that are themselves sparse. If the regularizer $\rho_r$ is chosen to discourage such solutions, it is possible to prove that the resulting objective function has no spurious local minimizers, and negative curvature at every non-minimizing critical point.

Equation \eqref{eqn:rpca} is just one model for matrix recovery from unreliable observations. Versions in which entire columns of $\mb Y$ are corrupted are also of interest for robust statistical estimation (see e.g., \cite{xu2010robust}), where they model outlying data vectors. Certain variants of this problem also inherit the geometry of factorization -- local minimizers are global, saddle points are generated by partial factorizations of the ground truth, and exhibit negative curvature in directions that introduce additional ground truth factors  \cite{lerman2018fast,maunu2019well,lerman2018overview}. It is also possible to formulate this version of the robust matrix recovery problem as one of finding a hyperplane that contains the majority of the datapoints. This dual viewpoint leads to nonconvex problems with a sign symmetry, which again have benign geometry under certain conditions on the input data \cite{tsakiris2017hyperplane,tsakiris2018dual,zhu2018dual,ding2021subspace}.

\section{Nonconvex Problems with Discrete Symmetry}
\label{sec:discrete}

In this section, we study nonconvex problems with discrete symmetry groups $\bb G$. Canonical examples include sparse dictionary learning (signed permutation symmetry) \cite{sun2017complete1,sun2017complete2,qu2019analysis,zhai2019complete}, sparse blind deconvolution (signed shift symmetry) \cite{zhang2017global,zhang2018structured,kuo2019dq,lau2019short,qu2019blind,li2018global}, tensor decomposition \cite{ge2015escaping,ge2017optimization} and clustering (permutation symmetry). Unlike the problem we discussed in \Cref{sec:rotational}, the problems of this type are not easily amenable to convexification; understanding nonconvex optimization landscapes becomes more critical. Design choices, such as the choice of objective function and constraints, also seem to play a critical role: many of the examples we review below are formulated as constrained optimization problems over compact manifolds such as spheres or orthogonal groups. We again begin by studying a very simple model problem: dictionary learning with one-sparse data. We extract several key intuitions for problems with discrete symmetries, and then examine how these intuitions carry over to less idealized (and more useful!) problem settings. 

\subsection{Minimal Example: Dictionary Learning with One-Sparse Data} 
\label{sec:toy_dl}

Similar to We introduce some basic intuitions through a model problem, which is a highly idealized version of {\em dictionary learning}. In this model problem, we observe a matrix $\mb Y$ which is the product of an orthogonal matrix $\mb A_0 \in O(m)$ (called a dictionary) and a matrix $\mb X_0 \in \reals^{m \times n}$ whose columns are one-sparse, i.e., each column of $\mb X_0$ has one nonzero entry:
\begin{equation}
\underset{\text{\bf \color{lava} data}}{\mb Y_{\color{white}0}} \;= \; \underset{\text{\bf \color{lava} orthogonal dictionary}}{\mb A_0} \quad \underset{\text{\bf \color{lava} 1-sparse coefficients}}{\mb X_0.} 
\end{equation}
This observation model exhibits a {\bf signed permutation symmetry} ($\bb G = \mr{SP}(n)$): for a given pair $(\mb A_0, \mb X_0)$, and any $\mb \Gamma \in \mr{SP}(n)$, the pair $(\mb A_0 \mb \Gamma, \mb \Gamma^* \mb X_0)$ also reproduces $\mb Y$. The goal is to recover $\mb A_0$ and $\mb X_0$, up to this signed permutation symmetry. A natural approach for recovering $\mb A_0$ is to search for an orthogonal matrix $\mb A$ such that $\mb A^* \mb Y$ is {\em as sparse as possible}: 
\begin{equation}
\min \; h( \mb A^* \mb Y ) \quad \text{s.t.} \quad \mb A \in O(m), \label{eqn:DL-ON}
\end{equation}
where $h( \mb X ) = \sum_{ij} h(\mb X_{ij})$ encourages sparsity. There are many possible choices for $h$ \cite{zhai2019complete,li2019nonsmooth,shen2020complete}; for concreteness, here we take $h$ to be the Huber function \cite{huber1992robust} as
\begin{equation}
h_\lambda(u) \;=\; \begin{cases} \lambda|u| - \lambda^2/2 & |u| > \lambda, \\ 
 u^2 / 2 & | u | \le \lambda. \end{cases} 
\end{equation}
The Huber loss can be viewed as a differentiable surrogate for the (sparsity promoting) $\ell^1$ norm. 

In \eqref{eqn:DL-ON}, finding the entire dictionary $\mb A_0 = \left[ \mb a_{01}, \dots, \mb a_{0m} \right]$ at once could still be quite challenging. Instead, an even simpler model problem is to solve for one column $\mb a_{0i}$ of $\mb A_0$ once at a time:
\begin{equation} \label{eqn:svias}
\min_{\mb a} \; h_{\lambda}\left( \mb a^* \mb Y \right) \quad \text{such that} \quad \mb a \in \bb S^{m-1}.
\end{equation}
Here, our goal is to recover a signed column $\pm \mb a_{0i}$ of the dictionary $\mb A_0$.\footnote{The entire dictionary can be recovered by solving a sequence of problems of this type; see \cite{spielman2013exact, sun2017complete1,sun2017complete2}.}
This problem asks us to minimize an $\ell^1$-like function over the sphere.\footnote{The problem \eqref{eqn:svias} can also be interpreted geometrically as searching for a sparse vector in the linear subspace $\mr{row}(\mb Y)$; see also \cite{qu2014finding, qu2020finding}.} 

To further simplify matters, we assume that the true dictionary $\mb A_0$ is an identity matrix. This does not change our geometric conclusions -- changing to another orthogonal $\mb A_0$ simply rotates the objective function. Similarly, because for this model problem each column of $\mb X_0$ has only one nonzero entry, we lose little generality in taking $\mb X_0 = \mb I$. 
With these idealizations, the problem simply becomes one of minimizing a sparsity surrogate over the sphere
\begin{equation}
\min \; \varphi(\mb a)\equiv  h_{\lambda} \left( \mb a \right) \quad \text{such that} \quad \mb a \in \bb S^{m-1}. \label{eqn:huber-sphere}
\end{equation}
Here, recovering a signed column of the true dictionary $\mb A_0 = \mb I$ corresponds to recovering one of the signed standard basis vectors $\pm \mb e_1, \dots, \pm \mb e_m$.

\begin{figure}[t]
\centerline{
\begin{tikzpicture}
\node at (-3.125,0) {\includegraphics[width=2in]{figures/huber_sphere.png}};
\node at (3.125,0) {\includegraphics[width=2in]{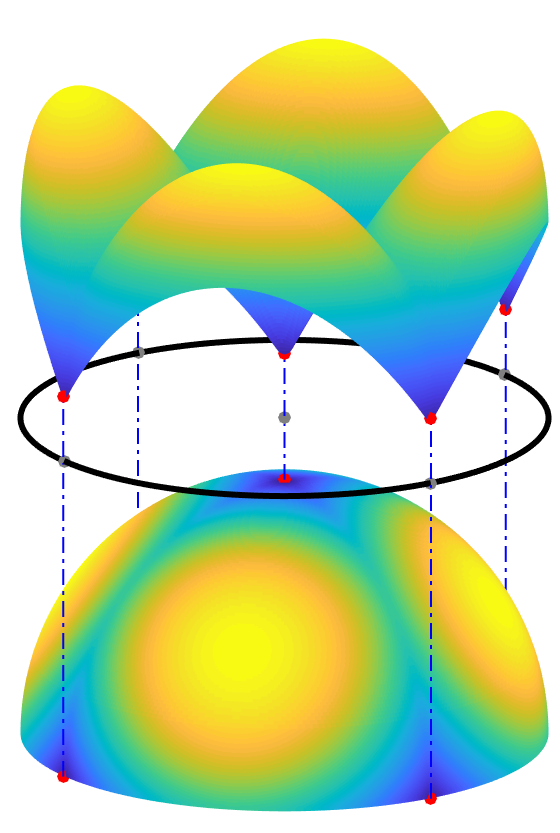}};
\end{tikzpicture}
}
\caption{{\bf A model problem with discrete symmetry.} 
Left: $h_{\lambda}(\mb u)$ as a function on the sphere $\bb S^{2}$. Local minimizers (red) are signed standard basis vectors $\pm \mb e_i$. These are the maximally sparse vectors on $\bb S^{2}$. Right: graph of $h_{\lambda}$; notice the strong negative curvature at points that are not sparse.} \label{fig:huber-sphere}
\end{figure}

\paragraph{Geometry of the Model Problem}
The 1-sparse dictionary learning model problem also exhibits a signed permutation symmetry: for any $\mb \Gamma \in \mr{SP}(m)$, $\varphi(\mb \Gamma \mb a) = \varphi(\mb a)$. The set of target solutions $\pm \mb e_1, \dots, \pm \mb e_m$ is also symmetric. \Cref{fig:huber-sphere} plots the objective function, and these target solutions, in a three dimensional example. Clearly, in this example, these target solutions are the only local minimizers. 

To study this phenomenon more formally, we need to understand the slope (gradient) and curvature (Hessian) of $\varphi$ as a function over the sphere $\bb S^{m-1}$. We develop these objects in an intuitive fashion. The sphere is a smooth manifold; its tangent space at a point $\mb a$ can be identified with the set of vectors $\mb a^\perp$ that are perpendicular to $\mb a$:
$$T_{\mb a} \bb S^{m-1} = \set{ \, \mb \delta \mid \mb a^* \mb \delta = 0 \, }.$$
The orthogonal projector onto the tangent space is simply given by $\mb P_{\mb a^\perp} = \mb I - \mb a \mb a^*$. The slope of $\varphi$ over the sphere (formally, the Riemannian gradient) is simply the component of the standard gradient that is tangent to the sphere:
\begin{equation}
\mr{grad}[\varphi](\mb a) = \mb P_{\mb a^\perp} \nabla \varphi(\mb a) 
\end{equation}
The curvature of $\varphi$ over the sphere is slightly more complicated than that of the Euclidean case. For a direction $\mb \delta \in T_{\mb a} \bb S^{m-1}$, the second derivative of $\varphi$ along the geodesic curve (great circle) $\gamma(t) = \exp_{\mb a}(t \mb \delta)$ is given by $\mb \delta^* \mr{Hess}[\varphi](\mb a) \mb \delta$, where $\mr{Hess}[\varphi]$ denotes the {\em Riemannian Hessian}
\begin{equation}
\mr{Hess}[\varphi](\mb a) \;=\; \mb P_{\mb a^\perp} \Bigl( \;\; \underset{\text{\bf \color{lava} curvature of $\varphi$}}{\nabla^2 \varphi(\mb a)} \,\quad - \quad \underset{\text{\bf \color{lava} curvature of the sphere}}{ \; \innerprod{ \nabla \varphi(\mb a) }{\mb a} \mb I} \quad \Bigl) \, \mb P_{\mb a^\perp}.
\end{equation}
This expression contains two terms. The first is the standard (Euclidean) Hessian $\nabla^2 \varphi$, which accounts for the curvature of the objective function $\varphi$. The second term accounts for the curvature of the sphere itself.\footnote{This expression can be derived in a simple way by letting $\| \mb \delta \|_2=1$, and calculating $\frac{d^2}{dt^2}\Bigr|_{t=0} \varphi\Bigl( \mb a \cos t + \mb \delta \sin t \Bigr)$.} Analogous to the case in the Euclidean space, critical points are characterized by Riemannian gradient $\mr{grad}[\varphi](\mb a) = \mb 0$; curvature can be studied through $\mr{Hess}[\varphi](\mb a)$. For more technical details, we refer readers to \cite{absil2009,boumal2020intromanifolds}.

To study the critical points, we begin by calculating the Euclidean gradient of $\varphi$:
\begin{align}
\nabla \varphi(\mb a) &= \lambda \,  \mr{sign}(\mb a) \odot  \indicator{|\mb a| > \lambda} \; + \; \mb a \odot  \indicator{| \mb a | \le \lambda},
\end{align}
where the Hadamard product $\odot$ denotes elementwise multiplication. Using this expression, we can show that the Riemannian gradient vanishes ($\mr{grad}[\varphi](\mb a) = \mb 0$) if and only if $\nabla \varphi(\mb a) \,\propto\, \mb a$ (here, $\propto$ denotes proportionality, i.e., $\exists s$ such that $\nabla \varphi(\mb a) = s \mb a$). This occurs whenever 
\begin{equation}
\mb a \; \propto \; \mr{sign}(\mb a). 
\end{equation}
We can therefore index critical points by the support $I$ and sign pattern $\mb \sigma$ of $\mb a$, writing $\mb a_{I,\mb \sigma}$. 
To understand which critical points are minimizers, we can study the Riemannian Hessian $\mr{Hess}[\varphi](\mb a)$. The Euclidean Hessian is 
$\nabla^2 \varphi(\mb a) = \indicator{| \mb a | \le \lambda}$; its Riemannian counterpart is 
\begin{equation}
\mr{Hess}[\varphi](\mb a_{I,\mb \sigma}) \;=\; \mb P_{\mb a_{I,\mb \sigma}^\perp} \left( \mb P_{|\mb a_{I,\sigma}| \le \lambda} - \lambda | I | \mb I \right) \mb P_{\mb a_{I,\mb \sigma}^\perp}.
\end{equation}
At critical points $\mb a_{I,\sigma}$ the Hessian exhibits $(|I| - 1)$ negative eigenvalues, and $n-|I|$ positive eigenvalues. Based on these calculations, we draw the following conclusions on the geometry of $\varphi$:
\begin{itemize}
\item {\bf \em Symmetric copies of the ground truth are minimizers}. Local minimizers are the signed standard basis vectors $\mb a = \pm \mb e_i$ with strictly positive Riemannian Hessians; the objective function is strongly convex in the vicinity of local minimizers.
\item {\bf \em Negative curvature in symmetry breaking directions}. Saddle points are balanced superpositions of target solutions: $\mb a_{I,\sigma} = \frac{1}{\sqrt{|I|}} \sum_{i \in I} \sigma_i \mb e_i$ for $I \subseteq \set{1, \dots, m }$ and signs $\sigma_i \in \set{\pm 1}$. There are negative curvatures in directions $\mb \delta \in \mr{span}( \{ \mb e_i \mid i \in I \} )$ that break the balance between target solutions.
\item {\bf \em Cascade of saddle points}. Saddle points are graded: points $\mb a_{I,\mb \sigma}$ with larger objective value have more directions of negative curvature. Moreover, similar to the examples discussed in the last section, the objective function exhibits a ``dispersive'' structure: downstream negative curvature directions are the image of upstream negative curvature directions under gradient flow. This means that negative curvature upstream helps to prevent local methods from stagnating near downstream saddle points. 
\end{itemize}
\noindent In the following subsections, we shall see how these basic phenomena recur in more practical nonconvex problems with discrete symmetries, including general dictionary learning (\Cref{sec:dl}), blind deconvolution (\Cref{sec:bd}), and others.



\subsection{Dictionary Learning}
\label{sec:dl}
The one-sparse dictionary learning problem is an extreme simplification of basic modern data processing problem: seeking a concise representation of data. As already precluded in \Cref{sec:toy_dl}, the goal of dictionary learning is to produce a sparse model for an observed dataset $\mb Y = \left[ \mb y_1, \dots, \mb y_p \right] \in \reals^{m \times p}$. Namely, we seek matrices $\mb A_0 \in \reals^{m \times n}$ and $\mb X_0 \in \reals^{n \times p}$ such that 
\begin{equation}
\mb Y \;\approx \; \underset{\text{\bf \color{lava} dictionary}}{\mb A_0} \quad \underset{\text{\bf \color{lava} sparse coefficients}}{\mb X_0} \label{eqn:DL-representation}
\end{equation}
 with $\mb X_0$ being as \emph{sparse} as possible. Sparsity is desirable for data compression, which facilitates tasks such as sensing, denoising, and superresolution, etc. \cite{wright2010sparse, elad2010sparse}
  
In the representation \eqref{eqn:DL-representation}, the data points $\mb y_j$ are approximated as superpositions $\mb y_j \approx \mb A_0 \mb x_{0j}$ of a few columns of the dictionary $\mb A_0 \in \bb R^{m \times n}$. Clearly, the size of the dictionary, $n$, has an impact on the accuracy, sparsity, and utility of this data representation. The appropriate dictionary size depends on application: for learning from a single image, a complete ($n = m$) dictionary may suffice, whereas for learning from a larger collection of images, an overcomplete ($n > m$) dictionary may be more appropriate \cite{murray2006learning, elad2006image, yang2010image}. Below, we discuss how our basic intuitions obtained from the orthogonal and one-sparse case in \Cref{sec:toy_dl}, can be carried over to each of these more realistic model problems. 

\paragraph{Complete Dictionary Learning} There are two basic issues in moving from the one-sparse dictionary learning problem to more general complete dictionary learning problems, in which $\mb A_0 \in \mathbb{R}^{m \times m}$ is some invertible matrix. First, the target dictionary $\mb A_0$ may not be orthogonal. Second, the columns of the coefficient matrix $\mb X_0$ are generally not one-sparse. For theoretical purposes, both of these issues can be addressed using probabilistic properties of $\mb X_0$. First, using the statistics of $\mb Y = \mb A_0 \mb X_0$ it is possible to reduce the problem of learning a general invertible $\mb A_0$ to one of learning an orthogonal matrix $\overline{\mb A} =\paren{\mb A_0 \mb A_0^*}^{-1/2} \mb A_0$. Concretely, if $\mb X_0$ is a sparse random matrix with independent symmetric entries \cite{sun2017complete1}, $\overline{\mb Y} = \paren{\mb Y \mb Y^*}^{-1/2} \mb Y \approx c \overline{\mb A} \mb X_0$ satisfies a sparse model with orthogonal dictionary $\overline{\mb A}$, where $c>0$ is a numerical constant. Similar to our discussion above, one can recover the columns of $\mb A_0$ by solving the optimization problem
\begin{align} \label{eqn:complete-analysis}
  \min \; \varphi(\mb a) \equiv h\left( \mb a^*\overline{\mb Y} \right) \quad\st\quad \mb a\in \bb S^{m-1}.
\end{align}
Although the columns of $\mb X_0$ are not one-sparse, when the number samples is large, this objective function retains all of the qualitative properties observed in the one-sparse problem, including local minimizers near symmetric solutions and saddle points near balanced superpositions of symmetric solutions, with negative curvature in symmetry breaking directions. The proofs of these properties rely heavily on probabilistic reasoning: one argues that the ``population'' objective function $\bb E_{\mb X_0} \brac{\varphi}$ has benign structure, and then argues that when the number $p$ of samples is large, gradients and Hessians of $\varphi$ are uniformly close to those of $\bb E \varphi$, and hence $\varphi$ has the same benign properties \cite{sun2017complete1,sun2017complete2}.

\paragraph{Overcomplete Dictionary Learning} In practice, {\em overcomplete} dictionaries, in which the number of dictionary atoms $n$ is larger than the signal dimension $m$, are often favored compared to complete dictionaries. Overcomplete dictionaries have greater expressive power, yielding sparser coefficient matrices $\mb X_0$. Our current theoretical understanding of the objective landscape associated with overcomplete dictionary learning is still developing. One suggestive result shows that when the dictionary is moderately overcomplete ($n \le 3m$), under appropriate technical hypotheses, a formulation based on maximizing the $\ell^4$ norm exhibits benign global geometry \cite{qu2019analysis}: again, every local minimizer is global and saddle points exhibit strict negative curvature.\footnote{When the dictionary is overcomplete, dictionary atoms $\mb a_i$ are correlated and $\mb a^* \overline{\mb Y}$ is no longer sparse, even if $\mb a$ is chosen as one of the atoms $\mb a_i$. Rather, at $\mb a = \mb a_i$, $\mb a^* \overline{\mb Y}$ is {\em spiky}, with a few large entries amongst many small ones. $\ell^4$ maximization is well-suited to encouraging this kind of spikiness. The most widely used practical dictionary learning algorithms are based on synthesis sparsity. Understanding the global geometry of this kind of formulation remains an important open problem} These results 
imply that overcomplete dictionary learning problems also exhibit benign global geometry. However, there are still a number of open questions around (i) the degree of overcompleteness $n / m$ that this structure can tolerate and (ii) the extent to which similar properties hold in more conventional {\em synthesis} dictionary learning formulations, in which one optimizes over both $\mb A$ and $\mb X$ simultaneously. 


%
%


\subsection{Sparse Blind Deconvolution}
\label{sec:bd}

Convolutional models arise in a wide range of problems in imaging and data analysis. The most basic convolutional data model expresses an observation $\mb y$ as the convolution of two signals $\mb a_0$ and $\mb x_0$. {\em Blind deconvolution} aims to recover $\mb a_0$ and $\mb x_0$ from the observation $\mb y$, up to certain intrinsic symmetries that we describe below. This problem is ill-posed in general -- there are infinitely many $(\mb a_0,\mb x_0)$ that convolve to produce $\mb y$. To make progress, some low dimensional priors about $\mb a_0$ and $\mb x_0$ are essential. Different priors yield different nonconvex optimization problems; in this section, we will focus on several variants of blind deconvolution with sparsity priors on $\mb x_0$, and then briefly mention other popular variants of blind deconvolution.

\subsection*{Short and Sparse Blind Deconvolution}

Analyzing signals comprised of repeated motifs is a common task in areas such as neuroscience, materials science, astronomy, and natural and scientific imaging \cite{starck2002deconvolution,pnevmatikakis2016simultaneous,cheung2020dictionary,lau2019short}. Such signals can be modeled as the {\em convolution} of a short motif $\mb a_0 \in \bb R^k$ and a sparse coefficient signal $\mb x_0 \in \bb R^m $, which encodes occurring locations of the motif in time/space. Mathematically, the observation $\mb y \in\R^m$ is the windowed\footnote{Rather than having complete access to the convolved signal (which could be infinitely long), we observe $m$ consecutive entries of it.} convolution of the short $\mb a_0$, which is supported on $k$ ($k\ll m$) consecutive entries, and the sparse $\mb x_0$:
\begin{align}
\label{eqn:linear_conv}
\mb  y =\proj{\mb a_0\ast\mb x_0}{m}.
\end{align}
Here, $\ast$ denotes linear convolution and $\proj{\cdot}{m}$ retains the entries supported on indices $0,\cdots,m-1$.

The inverse problem of recovering $\mb a_0$ and $\mb x_0$ from $\mb y$ is called {\em short and sparse} blind deconvolution (SaS-BD) \cite{zhang2017global, zhang2018structured, kuo2019dq}. The linear convolution $\ast$ exhibits a {\em signed shift symmetry}: 
\begin{align}
\mb a_0\ast\mb x_0=\alpha\shift{\mb a_0}{\tau}\ast\alpha^{-1}\shift{\mb x_0}{-\tau}.
\end{align}
Here $\alpha$ is some nonzero scalar and $\shift{\mb v}{\tau}$ denotes a shift of vector $\mb v$ by $\tau$ entries, i.e. $\shift{\mb v}{\tau}(i)=\mb v(i-\tau)$. As with the other nonconvex problems we have studied up to this point, we should expect this symmetry to play an critical role in shaping the landscape of optimization -- in particular, we would expect the {\em global} minimizers to be symmetric copies of the ground truth.\footnote{Notice that the scale and shift symmetries are intrinsic to the convolution operator in \eqref{eqn:linear_conv}. Although we focus on {\em sparse} deconvolution, these symmetries will persist in deconvolution with any shift-invariant structural model for $\mb a_0$ and $\mb x_0$. Moreover, as we will see below, they persist even in the presence of artificial symmetry-breaking mechanisms, in the sense that they still dictate the local minimizers.}

\paragraph{Symmetry Breaking?} However, there is a wrinkle: in order to obtain a finite dimensional optimization problem, one typically constrains the length-$k$ signal $\mb a_0$ to be supported on $\{ 0, \dots, k-1 \}$. This constraint appears to remove the shift symmetry: now only a scaled version $(\alpha \mb a_0, \alpha^{-1} \mb x_0)$ of the truth exactly reproduces the observation. Perhaps surprisingly, even with this constraint, symmetry {\em still} shapes the landscape of optimization. However, instead of dictating the global minimizers, in constrained formulations, symmetry dictates the {\em local} minimizers. The reason is simple: a shift of $\mb a_0$ by $\tau$ samples is not supported on $\{ 0, \dots, k-1 \}$, and hence is not feasible. Nonetheless, its truncation to $\{0,\dots, k-1\}$ {\em is} feasible, and still approximates $\mb y$: 
\begin{align}
\mb y\approx\proj{\shift{\mb a_0}{\tau}}{k}\ast \shift{\mb x_0}{-\tau}.
\end{align}
Because this approximation is not perfect, truncated shifts are \emph{not} global minimizers, while they are very close to {\em local} minimizers \cite{zhang2017global, zhang2018structured}. These points have suboptimal objective value and do not exactly reproduce $(\mb a_0,\mb x_0)$. Despite this, the optimization landscape is still sufficiently benign\footnote{In particular, there is negative curvature in symmetry breaking directions.} that it is possible to exactly recover $(\mb a_0, \mb x_0)$ with efficient methods -- one can, e.g., first find a local minimizer that is close to a truncated shift of $\mb a_0$, and then refine it to exactly recover $\mb a_0$ \cite{zhang2017global,kuo2019dq}. 

This problem illustrates how hard it is to avoid symmetry in studying deconvolution problems: even with an explicit symmetry breaking constraint, symmetry still shapes the landscape of optimization! The main motivation for studying this more complicated deconvolution model is its applicability. Giving formulations that better respect the symmetry structure, and hence have no spurious local minimizers, is an important open problem. 

\subsection*{Multi-channel Sparse Blind Deconvolution}
The problem of  multi-channel sparse (MCS) blind deconvolution assumes access to \emph{multiple} observations $\mb y_i=  \proj{\mb a_0\ast\mb x_i}{m}\;(1\leq i\leq n)$ generated from $n$ circular convolution\footnote{The linear convolution is a better model for many practical application. Despite this, there is no loss of generality as any statements about cyclic convolution, which can easily be carried over to linear convolution; by zero-padding $\mb x_i$ appropriately, one can always rewrite a linear convolution as a cyclic convolution.}  of $\mb a_0\in\reals^m$ and distinct sparse signals $\mb x_i \in \bb R^m$ \cite{li2018global,qu2019blind,qu2019analysis,shi2019manifold}. Here, shift symmetry becomes a {\em cyclic} shift symmetry: there exist $m$ equivalent solutions corresponding to $m$ different cyclic shifts. The resulting optimization landscape exhibits similar characteristics to that of complete dictionary learning, which we have described in Section \ref{sec:toy_dl} and Figure \ref{fig:huber-sphere}. In particular, any local minimizer is a scaled cyclic shift of the ground truth \cite{li2018global, qu2019blind,shi2019manifold}.

\subsection*{Geometry of Sparse Blind Deconvolution}
Despite the technical differences of the convolution operator in MCS and SaS blind deconvolution problems, their optimization landscapes share the following key common phenomena:
\begin{itemize}

\item {\bf \em Symmetric copies of the ground truth are minimizers}. The local minimizers are either a cyclic shifted or shifted truncation of the ground truth under conditions. Both can be viewed as a result of the inherent shift symmetry. 

\item {\bf \em Negative curvature in symmetry breaking directions}. Near saddle points, there is negative curvature in the direction of any particular (truncated) shifted copy of the ground truth, and the objective value decreases by moving towards this symmetry breaking direction.

\item {\bf \em Cascade of saddle points}. The saddle points are approximately balanced superpositions of several shifts of the ground truth. The more shifts participate, the larger the objective value and the more negative eigenvalues the Hessian exhibits.  
%
\end{itemize}

\subsection*{Other Blind Deconvolution Variants}
\noindent {\bf Subspace Blind Deconvolution} is another widely studied variant of blind deconvolution that leverages a low dimensional model for the pair $(\mb a_0,\mb x_0)$. In this variant,  $\mb a_0$ and $\mb x_0$ are assumed to lie on known low-dimensional subspaces \cite{ahmed2014blind}. This problem can be cast as a rank-one matrix recovery problem, which exhibits a similar geometry to the problems studied in Section \ref{sec:rotational}. \newline

\noindent {\bf Convolutional Dictionary Learning} extends the basic convolution model 
\eqref{eqn:linear_conv} by allowing for multiple basic motifs $\mb a_1, \dots, \mb a_N$  \cite{garcia2018convolutional}. More precisely, we observe one or more signals of the form $\mb y = \sum_{i=1}^N \mb a_i\ast\mb x_i,$ and the goal is to recover all the $\mb a_i$ and $\mb x_i$.
In addition to the symmetries inherited from the convolution operator, this problem processes an additional {\em permutation symmetry}: permuting the index $i$ does not change the approximation to $\mb y$. Despite this additional complexity, empirically local minimizers remain symmetric copies of the ground truth \cite{zhang2017global,lau2019short}; under certain technical hypotheses, one can prove that natural first order algorithms always recover one such symmetric copy \cite{qu2019analysis}.

\subsection{Other Nonconvex Problems with Discrete Symmetry}
\label{sec:other_dis}

\paragraph{Symmetric Tensor Decomposition} Tensors can be regarded as high dimensional generalizations of matrices. Tensor decomposition problems find many applications in statistics, data science, and machine learning \cite{kolda2009tensor, anandkumar2014tensor, sidiropoulos2017tensor,MAL-057}. Although we can usually generalize algebraic notions from matrices to tensors, their counterpart in tensors are often not as well-behaved or easy to compute \cite{kolda2009tensor}. In fact, many natural tensor problems are NP-hard in the worst-case \cite{hillar2013most}.

Nonetheless, recent results suggest that certain appealing special cases of tensor decomposition are tractable \cite{anandkumar2014tensor,ge2015escaping,MAL-057}. This is especially true for orthogonal tensor decomposition, where the task is to decompose a $p$-th order symmetric tensor (e.g., $p = 3,4,\cdots$) into these orthogonal components. More specifically, an orthogonal tensor $\mc T$ with $r$ components can be presented in the following form
\begin{align}\label{eqn:tensor-orth}
   \mc T \;=\; \sum_{ k=1 }^r\; \mb a_k^{ \otimes p }, 	\qquad r \leq n,
\end{align}
with $ \Brac{ \mb a_k }_{k=1}^r $ are a collection of orthogonal vectors, and $\mb a^{\otimes p}$ denotes the $p$-way outer product of a vector $\mb a$. The orthogonal tensor decomposition shares many similarities with the other nonconvex problems with discrete symmetry discussed above:
\begin{itemize}
	\item the problem exhibits a \emph{signed permutation symmetry} which is similar to dictionary learning: given $\mc T$ we can only hope to recover the orthogonal components $ \Brac{ \mb a_k }_{k=1}^r $ up to an order permutation;
	\item whenever $p$ is a number of even order, as shown in \Cref{fig:discrete-symmetries}, a natural nonconvex formulation
	\begin{align}\label{eqn:tensor-opt}
	    \min_{\mb x \in \bb S^{n-1} } \; -\mc T(\mb x,\cdots,\mb x) \;=\; -\norm{ \mb A^* \mb x }{p}^p \quad \text{with}\quad \mb A\;=\; \begin{bmatrix}
	    	\mb a_1 & \cdots & \mb a_r
	    \end{bmatrix}
	\end{align}
	manifests a similar optimization landscape, for which every local minimizer is close to one of the signed orthogonal components and other critical points exhibit strict negative curvature. 
\end{itemize}
These results have inspired further endeavors beyond orthogonal tensors \cite{qu2019analysis,sanjabi2019does,ge2017optimization}. One particular case of interest is decomposing a symmetric tensor $\mc T$ in \eqref{eqn:tensor-orth} with $r > n$ and nonorthogonal $ \Brac{ \mb a_k }_{k=1}^r $, which is often referred as \emph{overcomplete} tensor decomposition. In particular, when $p=4$, $ r \in \mc O(n^{1.5})$ and $ \Brac{ \mb a_k }_{k=1}^r $ are i.i.d.\ Gaussian, \cite{ge2017optimization} shows that \eqref{eqn:tensor-opt} has no bad local minimizer over a level set whose measure geometrically shrinks w.r.t.\ the problem dimension; for $p=4$, $ r <3n$, and incoherent $ \Brac{ \mb a_k }_{k=1}^r $, \cite{qu2019analysis} presented a global analysis for overcomplete tensor decomposition, disclosing its connection to overcomplete dictionary learning. Nonetheless, these results are still far from providing a complete understanding of overcomplete tensor decomposition. One interesting question remains largely open is when bad local minimizers exist for large rank $r\gg n$ in the nonorthogonal case.


\paragraph{Clustering} Clustering is one of the most fundamental problems in unsupervised learning. This problem possesses a {\em permutation symmetry}: one can generate equivalent clusters by permuting the indices for cluster centers. Popular nonconvex algorithms include the Lloyd algorithm and variants of Expectation-Maximization. Despite the broad applications and empirical success of these methods, few theoretical guarantees have been obtained until recently. The problem of demixing two balanced, identical data clusters manifests global convergence to (a symmetric copy of) the ground truth \cite{balakrishnan2017statistical, xu2016global, daskalakis2016ten, qian2019global, kwon2019global}. We see similar geometric properties hold here: {\em symmetric copies of the ground truth are minimizers} and {\em saddle points exhibit directions of strict negative curvature}. Moreover, the saddle points are also located at balanced superpositions of local minimizers. Sometimes, these saddle points may contain redundant cluster estimates. In this case, the redundant cluster estimates can be interpreted as an under-parametrized solution (with a smaller $k$ specified).

However, in general, clustering problems with more than two clusters, local minimizers provably exist \cite{dasgupta2007probabilistic, jin2016local}. When the clusters are sufficiently separated, these local minimizers possess characteristic structures \cite{qian2020structures}: they correspond to imbalanced segmentations of the data, in which a subset of the true clusters are optimally under-segmented and another subset is optimally over-segmented. 


\paragraph{Fourier Phase Retrieval}
How to efficiently solve {\em Fourier} phase retrieval is a crucial problem in scientific imaging. In this problem, the goal is to recover $\mb x_0$ from observation $\mb y = \abs{\mc F(\mb x_0)}$.
Apart from the rotational (phase) symmetry, the problem of Fourier phase retrieval manifests two additional symmetries:\footnote{When $\mb x$ is one dimensional, the problem becomes even more pessimistic --- there exist multiple one dimensional signals with the same Fourier magnitude, but not related by an obvious symmetry.} {\em (cyclic)-shift symmetry} $\abs{\mc F(\mb x)}= \abs{\mc F(\shift{\mb x}{\tau})}$ and {\em conjugate inversion symmetry} $\abs{\mc F(\mb x)} = \abs{\mc F(\check{\mb x})}$, where $\check{\mb x}(n) = \bar{\mb x}(-n)$ \cite{bendory2017fourier}. This complicated symmetry structure is reflected in a complicated optimization landscape, which is challenging to study analytically. Many basic problems in the algorithmic theory of Fourier phase retrieval remain open.


\section{Discussion}
\label{sec:discussion}

Finally, we conclude this review by pointing several important future directions. We start with a discussion on training deep neural networks. We show how the symmetry and geometry can play crucial roles in analyzing some interesting phenomena in deep learning. Second, we conclude by touching some important generic questions in nocnonvex optimization, such as relationship to convexification, development of more unified theory, and more efficient optimization methods.

\subsection{Symmetry \& Geometry in Training Deep Neural Networks}\label{subsec:nc-dnn}

In the past decade, the revival of deep neural networks (DNN) has led to dramatic success in numerous applications ranging from computer vision, to natural language processing, to scientific discovery and beyond \cite{krizhevsky2012imagenet,lecun2015deep,goodfellow2016deep,senior2020improved}.  However, because most neural network learning problems are highly nonconvex, to date a deep network is still viewed as a \emph{black-box} and the practice of deep networks has been shrouded with mystery. 

Taking the multi-class (say $K$ classes) classification problem for an example, the goal of deep network training is to learn a classifier $\mb W$ and deep hierarchical representation (or feature) $\mb h(\mb x)$, such that the output $\psi(\mb x) = \mb W \mb h(\mb x)$ of the network fits the input $\mb x$ to the corresponding (one-hot) training label $\mb y$. In the vanilla form, a $L$-layer neural network can be written as
\begin{align}\label{eq:func-NN}
    \psi_{\mb \Theta}(\mb x_{i,k}) \;=\;  \underset{ \text{\bf \color{lava} linear classifier}\;\mb W }{ \mb W_L} \; \cdot  \; \underset{\text{ \bf \color{lava} feature}\;\; \mb h_{i,k}\;=\; \phi_{\mb \theta}(\mb x_{i,k})}{\sigma\paren{ \mb W_{L-1} \cdots \sigma \paren{\mb W_1 \mb x_{i,k} + \mb b_1} + \mb b_{L-1} }} \; + \; \mb b_L,
\end{align}
where each layer is composed of an affine transformation, represented by some weight matrix $\mb W_\ell$, and bias $\mb b_\ell$, followed by a nonlinear activation function $\sigma(\cdot)$, and $\mb \Theta = \Brac{ \mb W_\ell, \mb b_\ell }_{\ell=1}^L$ denotes the network parameters. To learn those weight parameters $\mb \Theta$, one typically minimizes an empirical risk over training samples of the following form
\begin{align}\label{eqn:dl-loss}
    \min_{\mb \Theta} \; \sum_{k=1}^K \sum_{i=1}^{n} L \paren{ \psi_{\mb \Theta}(\mb x_{k,i}),\mb y_k } \;+\; \frac{\lambda}{2} \norm{\mb \Theta}{F}^2,
\end{align}
where $\mb y_k \in \bb R^K $ is a one-hot vector with only the $k$th entry equal to unity ($1\leq k \leq K$), $n$ are the numbers of training samples in each class,\footnote{For simplicity, we assume balanced class, that each class has the same number of training samples.} and $\lambda>0$ is the regularization parameter (or weight decay penalty), and $L(\cdot,\cdot)$ is a data fidelity term (e.g., cross entropy loss). As we observe from \eqref{eq:func-NN}, deep neural networks typically exhibit complicated symmetries which include compositions of permutations, and nonlinear interactions between layers, so that the training loss \eqref{eqn:dl-loss} is highly nonconvex. This formulation can be naturally generalized from classification to regression problems. Studying the nonconvex optimization landscape in neural network training has been an active research area recently; see \cite{vidal2017mathematics,sun2020optimization,sun2020global,ma2020towards,berner2021modern,fang2021mathematical} for contemporary surveys.

\paragraph{Study of Multi-layer Linear Neural Network} To simplify the problem, people considered  {\em linear} neural networks by removing all nonlinear activation $\sigma(\cdot)$ in \eqref{eq:func-NN} as a more approachable object of theoretical investigation. This model exhibits rotational symmetries at each layer. Using similar considerations to those described in \Cref{sec:lowrank} for low-rank matrix factorization,
 the line of work \cite{baldi1989neural,kawaguchi2016deep,nouiehed2018learning,zhu2019global,yun2018global} studied the optimization landscape for linear two-layer networks and proved that the associated training loss is a strict saddle function.  As with matrix factorization, critical points of natural optimization models correspond to ``underfactorizations''. Additionally, some related works study the learning dynamics of gradient descent \cite{arora2018convergence,arora2019implicit}. For deeper linear networks, however, in contrast to matrix factorization, this problem does possess ``flat'' saddle points at which the Hessian has no negative eigenvalues -- this is the result of the compound effect of symmetries at multiple layers, but there are no spurious local minima \cite{kawaguchi2016deep,laurent2018deep}. Nonetheless, because the power of deep neural network originates from its nonlinearity, the removal of the nonlinear interaction makes linear deep neural networks far from practice.

\paragraph{Study of Nonlinear Neural Network} 
 When the nonlinear activations in \eqref{eq:func-NN} are considered, the problem has more complicated symmetry groups than the problems described above. For example, natural objective functions associated with fitting a fully connected neural network are invariant under simultaneous permutations of the features at {\em each} layer.  We currently lack tools for reasoning about the global geometry of such problems. However, progress has been made on certain special cases: for example, certain problems associated with fitting shallow networks share similar geometry to tensor decomposition \cite{janzamin2015beating, mondelli2018connection}. With varying technical assumptions, all local solutions have been shown to be global in a 1-layer neural network \cite{haeffele2017global,feizi2017porcupine,ge2017learning,gao2018learning,soltanolkotabi2018theoretical}. However, general deep nonlinear neural networks can exhibit flat saddles and spurious local minimizers for all the weight parameters \cite{safran2017spurious,vidal2017mathematics,yun2018small}, but such local minima may be eliminated, or the number can be significantly reduced, in the over-parameterization regime \cite{safran2018spurious,liang2018adding}. Additionally, the work \cite{haeffele2015global} proved that certain local minima (having an all-zero ``slice'') are also global solutions, but the analysis is crucially dependent on the sufficient condition of an all-zero slice in the weights, which is insufficient to characterize the landscape properties. 


\paragraph{Investigation Under Simplified Unconstrained Feature Model}
To avoid the challenges on dealing with nonlinear interactions across layers, recent works simplifies the study of deep network \eqref{eq:func-NN} by assuming a certain \emph{unconstrained feature model} (UFM) \cite{mixon2020neural,lu2020neural,weinan2020emergence,fang2021exploring,graf2021dissecting,ergen2021revealing,zhu2021geometric,ji2022an,zhou2022optimization} and investigate the structure of last-layer representations -- by making an assumption that the representation $\mb h_{i,k}$ of each input $\mb x_{i,k}$ in \eqref{eq:func-NN} is a \emph{free} optimization variable, we can avoid the nonlinear interactions between layers and reduce the problem similar to matrix factorization with \emph{only} the rotational symmetry. The underlying motivation behind UFM is that modern neural networks are highly overparameterized to approximate any continuous function \cite{cybenko1989approximation, hornik1991approximation,lu2017expressive,shaham2018provable}. Thus, based upon the UFM, recent work \cite{zhu2021geometric,zhou2022optimization} reduced \eqref{eqn:dl-loss} to much simpler optimization problems such as the following
\begin{align}\label{eq:obj}
     \min_{\mb W , \mb H,\mb b  } f(\mb W,\mb H,\mb b):=  \frac{1}{Kn} \sum_{k=1}^K \sum_{i=1}^{n} L \paren{ \mb W \mb h_{k,i} + \mb b, \mb y_k } + \frac{\lambda_{\mb W} }{2} \norm{\mb W}{F}^2 + \frac{\lambda_{\mb H} }{2} \norm{\mb H}{F}^2 + \frac{\lambda_{\mb b} }{2} \norm{\mb b}{2}^2 ,
\end{align}
with $\mb W \in \bb R^{ K \times d}$, $\mb H = \begin{bmatrix}\mb h_{1,1} \cdots \mb h_{K,n} \end{bmatrix}\in \bb R^{d \times N}$ (here, we denote $N = nK$), $\mb b \in \bb R^K$, and $\lambda_{\mb W},\lambda_{\mb H},\lambda_{\mb b}>0$ are the penalty parameters. The difference of \eqref{eq:obj} from \eqref{eqn:dl-loss} is that the regularization is on $\mb W$ and $\mb H$ instead of all network parameters $\mb \Theta$. Recent work \cite{zhu2021geometric,ji2022an,zhou2022optimization} showed that the nonconvex objective \eqref{eq:obj} exhibits similar symmetry and geometric structure as the low-rank matrix recovery that we discussed in \Cref{sec:lowrank}. While all saddle points of \eqref{eq:obj} are also strict saddles whose Hessian exhibit negative curvature directions, the minor differences lie in the global solutions. Because of the particular structure of the one-hot encoders $\mb y_k$, the only global solutions of \eqref{eq:obj} are more structured -- they satisfies certain properties termed \emph{Neural Collapse} (\NC) \cite{papyan2020prevalence,han2022neural}. This phenomenon has been empirically revealed as well for training practical deep networks \cite{papyan2020prevalence,han2022neural}, that (i) the class means and the last-layer classifiers all collapse to the vertices of a Simplex Equiangular Tight Frame (ETF) up to scaling, and (ii) cross-example within-class variability of last-layer representations collapses to zero. Under the UFM assumption, recent works showed that the global optimality and \NC\ happen for a variety of loss function $L(\cdot)$, such as cross-entropy \cite{papyan2020prevalence,lu2020neural,zhu2021geometric,fang2021exploring,ji2022an}, mean-squared error \cite{mixon2020neural,han2022neural,zhou2022optimization,tirer2022extended}, and supervised contrastive loss \cite{graf2021dissecting}. Moreover, the simplified analysis based upon UFM not only characterizes the features that are learned in the last layer, but also explains why they can be efficiently optimized. The study of symmetry and geometry provides theoretical support for empirical observations in practical deep network architectures.

Although it sounds plausible, however, it should be noted that the presented global landscape analysis in \cite{zhu2021geometric,ji2022an,zhou2022optimization} is \emph{only} with respective to the last-layer feature $\mb H$ and classifier $\mb W$, \emph{not} the network parameters $\mb \Theta$. For deep network training, it should be noted that spurious local minimia \emph{do exist} for $\mb \Theta$ \cite{safran2018spurious,sun2019optimization}, and the training algorithm often has implicit bias towards certain low-dimensional solutions \cite{soudry2018implicit,arora2019implicit,you2020robust,liu2022robust}.

\subsection{Methodological Points \& Future Directions}

This work has reviewed recent advances in provable nonconvex methods for signal processing and machine learning, through the lens of symmetry and geometry. It is an exciting time to work on both the theory and practice of nonconvex optimization.  For complementary perspectives on the area, we refer interested readers to other recent review papers \cite{jain2017non,sun2019link,chi2019nonconvex,qu2020finding}. In the following, we close by discussing several general methodological points and general directions for future work.

\paragraph{Convexification} In the past decades, convex relaxation has been demonstrated a powerful tool for solving nonconvex problems such as sparse recovery \cite{candes2007sparsity,candes2006stable} low-rank matrix completion \cite{candes2010power,candes2009exact,candes2011robust}, etc. For these problems, convex relaxation achieves near-optimal sample complexity.  Which nonconvex problems are amenable to convex relaxation? There are general results that suggest that {\em unimodal} functions (i.e., functions with one local minimizer) on convex sets can be convexified, by endowing the space with an appropriate geometry \cite{rapcsak1993nonlinear}.\footnote{These are existence results; their direct implications for efficient computation are limited, since they apply to NP-hard problems. It is also worth noting that many of our discrete symmetric problems in Section \ref{sec:discrete} are formulated over compact manifolds such as $\bb S^{n-1}$; the only continuous geodesically convex function on a compact Riemannian manifold is a constant \cite{bishop1969manifolds,yau1974non}.} The symmetric problems encountered in this survey are not unimodal. The degree to which they are amenable to convex relaxation varies substantially:
\begin{itemize}
	\item \textbf{\emph{Problems with rotational symmetry}.} Many problems with rotational symmetry {\em can} be convexified by lifting to a higher dimensional space \cite{candes2009exact,candes2011robust,candes2013phaselift}, e.g., by replacing the factor $\mb U$ with a matrix valued variable $\mb X = \mb U \mb U^*$. This collapses the $O(r)$ symmetry; the resulting problems can often be converted to semidefinite programs and solved globally. Typically, nonconvex formulations are still preferred in practice, due to their scalability to large datasets. Section \ref{sec:rotational} and the references therein describe alternative geometric principles that help to explain the success of these methods. 	
	\item \textbf{\emph{Problems with discrete symmetry}.} Most of the discrete symmetric problems described in Section \ref{sec:discrete} do not admit simple convex relaxations. For example, complete dictionary learning can be reduced to a sequence of linear programs \cite{spielman2013exact}, but only in the highly sparse case, in which the target sparse representation has $O(\sqrt{n})$ nonzero entries per length-$n$ data vector. These limitations are attributable in part to the more complicated discrete symmetry structure. Natural ideas, such as taking a quotient by the symmetry group, encounter obstacles at both the conceptual and implementation levels. One general methodology which {\em does} meet with success in this setting is sum-of-squares relaxation, which for variants of dictionary learning and tensor decomposition leads to quasipolynomial or even polynomial-time algorithms \cite{barak2015dictionary}.  
\end{itemize}


\paragraph{Disciplined Formulations and Analysis} Our understanding of nonconvex optimization is still far from satisfactory -- analyses are delicate, case-by-case, and pertain to problems with elementary symmetry (e.g., rotation or permutation) and simple constraints (e.g., the sphere).
\begin{itemize}
    \item \textbf{\emph{A unified theory for nonconvex optimization}.} Analogous to the study of convex functions \cite{boyd2004convex}, there is a pressing need for simpler analytic tools, to identify and generalize benign properties for new nonconvex problems, despite some recent endeavors \cite{qu2019analysis,li2019nonsmooth} of identifying general conditions and operations preserving benign geometric structures. 
    \item \textbf{\emph{Coping with complicated symmetries and constraints}.} Practical nonconvex problems often involve \emph{multiple symmetries} (e.g., Fourier phase retrieval and deep neural networks) and/or \emph{complicated manifolds} (e.g., the Stefiel manifold \cite{hu2019brief}). We need better technical tools to understand the impact of compound symmetries (especially compound discrete symmetries) on the optimization landscape, despite some steps in this directions~\cite{li2019nonsmooth,hu2019brief,zhai2019complete}.
    \item \textbf{\emph{Dealing with nonsmoothness}.} In many scenarios we encounter nonconvex problems with \emph{nonsmooth} formulations \cite{davis2018subgradient,davis2018graphical,li2020nonconvex,li2019nonsmooth,bai2019subgradient,zhu2018dual,charisopoulos2019composite,charisopoulos2019low}, for better promoting solution sparsity or robustness. However, most of our current analysis is local \cite{charisopoulos2019composite,li2019nonsmooth}, and (subgradient) optimization \cite{li2019nonsmooth,bai2019subgradient,zhu2018dual,ding2021rank} could be slow to converge. Attempts to obtain global analyses and fast optimization methods might benefit from more sophisticated tools from variational analysis \cite{rockafellar2009variational} and development of efficient $2$nd-order methods \cite{duchi2019solving}.    
\end{itemize}

\paragraph{Efficient First-Order Algorithms} 

In this paper, we have described families of symmetric nonconvex optimization problems with benign global geometry: local minimizers are global and saddle points exhibit strict negative curvature. Although we have not emphasized algorithmic aspects of these problems, this geometric structure {\em does} have strong implications for computation -- a variety of methods the key is leveraging negative curvature to efficiently obtain minimizers. One class of methods explicitly models negative curvature, e.g., using a second-order approximation to the objective function. Methods in this class include trust-region methods \cite{conn2000trust}, cubic regularization \cite{nesterov2006cubic}, and curvilinear search  \cite{goldfarb1980curvilinear}. These methods can be challenging to scale to very large problems, since they typically require computation and storage of the Hessian. It is also possible to leverage negative curvature using more scalable first-order methods such as gradient descent. In the vicinity of a saddle point, the gradient method essentially performs a power iteration that moves in directions of negative curvature. Although this scheme {\em can} stagnate at or near saddle points, it is possible to guarantee efficient escape by perturbing the iterates with an appropriate amount of random noise \cite{ge2015escaping,jin2017escape,jin2017accelerated,criscitiello2019efficiently,sun2019escaping}. 

The methods described above are efficient across the broad class of {\em strict saddle functions} \cite{ge2015escaping,sun2015nonconvex}, i.e., functions whose saddle points all have directions of strict negative curvature. This is a worst case performance guarantee. Perhaps surprisingly, the simplest and most widely used first order method, gradient descent, is not efficient for worst case strict saddle functions: although randomly initialized gradient descent {\em does} obtain a minimizer with probability one \cite{lee2016gradient,lee2019first}, for certain strict saddle functions it can take time exponential in dimension \cite{du2017gradient}. These challenging functions have a large numbers of saddle points, which are conspicuously arranged such that upstream negative curvature directions align with {\em positive} curvature directions for downstream saddle points. 

This worst case behavior is in some sense the opposite of what is observed in the type of highly symmetric functions studied here: functions encountered in generalized phase retrieval \cite{chen2018gradient}, dictionary learning \cite{gilboa2018efficient}, deconvolution\cite{qu2019blind,shi2019manifold}, etc., exhibit a global negative curvature structure (\cite{gilboa2018efficient} and Appendix \ref{app:dispersive}, in which upstream negative curvature directions align with {\em negative} curvature directions of downstream saddle points. In this situation, randomly initialized gradient descent is efficient. This points to another gap between naturally occurring nonconvex optimization problems and their worst case counterparts. There is substantial room for future work in this direction.

\section*{Acknowledgement} This work was supported by the grants NSF CCF 1527809, NSF CCF 1740833, NSF CCF 1733857, and NSF IIS 1546411. YZ acknowledges support from Electrical and Computer Engineering Department at Rutgers University. QQ acknowledges past support from Microsoft Ph.D. fellowship, Moore-Sloan fellowship, and the grant NSF DMS 2009752 for conducting part of the research, and also acknowledges current funding support from U-M START \& PODS grants, NSF CAREER CCF 2143904, NSF CCF 2212066, NSF CCF 2212326, and ONR N00014-22-1-2529. We also thank our colleagues Jun Sun (University of Minnesota), Yuxin Chen (University of Pennsylvania), Yi Ma (UC Berkeley), Zhihui Zhu (University of Denver), Sam Buchanan (Columbia University), Han-wen Kuo (Google) for discussion and contributions on related works. 

\newpage 

{\small
\bibliographystyle{ieeetr}
\bibliography{ncvx,deconv,pr,lowrank,dictl,deepl,tensor,clustering}
}

\newpage
\appendix 

%

\section{Critical Points of Low Rank Matrix Factorization}
\label{app:factorization}

In this appendix, we give a more detailed accounting of the critical points of two model matrix factorization problems.

\paragraph{Symmetric Low Rank Matrix Factorization}

We begin by considering the symmetric factorization problem \eqref{eqn:symmetric-factorization}. This is a nonconvex optimization problem, with orthogonal symmetry $\mb U \equiv \mb U \mb \Gamma$. As we will see, its critical points can be described in terms of the eigendecomposition of the symmetric matrix $\mb X_0$. Here, $\mb X_0$ has a complete orthonormal basis of eigenvectors $\mb \xi_1, \dots \mb \xi_n$, with corresponding nonnegative eigenvalues $\lambda_1 > \lambda_2 > \dots > \lambda_r > \lambda_{r+1} = \dots = \lambda_n = 0$.\footnote{For simplicity, we assume that the nonzero eigenvalues are distinct. Problems with repeated eigenvalues exhibit a similar structure, with minor modifications.} In other words, we can write
\begin{equation}
\mb X_0 = \sum_i \lambda_i \mb \xi_i \mb \xi_i^*.
\end{equation}
Using properties of the eigenvectors, it is not difficult to show that every optimal factorization $\mb X_0 = \mb U \mb U^*$ can be written as
\begin{equation} \label{eqn:opt-factorization}
\mb U = \left[ \lambda_1^{1/2} \mb \xi_1  \, \mid \, \dots \, \mid \,  \lambda_r^{1/2} \mb \xi_r  \right] \mb \Gamma,
\end{equation}
for some orthogonal matrix $\mb \Gamma \in O(r)$. 
By setting $\nabla \varphi = \mb 0$, we can obtain the following characterization of critical points: $\mb U$ is a critical point if and only if it can be written as 
\begin{equation}
\mb U = \left[ \mb \phi_1 \mid \mb \phi_2 \mid \dots \mid \mb \phi_r \right]  \mb \Gamma, \qquad \mb \Gamma \in O(r),
\end{equation}
where the columns $\mb \phi_j\;(1\leq j \leq r',\; r'\leq r)$ are generated by appropriately scaling some orthogonal eigenvector of $\mb X_0$, so that
\begin{equation}
\mb \phi_1 = \lambda_{i_1}^{1/2} \mb \xi_{i_1}, \; \dots, \; \mb \phi_{r'} = \lambda_{i_{r'}}^{1/2} \mb \xi_{i_{r'}},
\end{equation}
with the indices $\Brac{ i_1,i_2,\cdots,i_{r'} } \subseteq [r]$, 
and setting any remaining $\mb \phi_\ell$ to $\mb 0$. 
In words, equivalence classes of critical points are generated by selecting subsets of the eigenvectors of $\mb X_0$. Selecting the $r$ leading eigenvectors, as in \eqref{eqn:opt-factorization}, gives a global minimizer. The curvature at other critical points can be studied through the Hessian
\begin{align}
\nabla^2 \varphi [ \mb W, \mb W ] = \tfrac{1}{2} \norm{ \mb U \mb W^* + \mb W \mb U^* }{F}^2 + \innerprod{ \mb U\mb U^* - \mb Y }{ \mb W\mb W^*}.
\label{eqn:hessian-matrix}
\end{align}
Evaluating this at critical points, and using orthogonality of the eigenvectors $\mb \xi_i$, we can observe that:
\begin{itemize}
\item {\em Saddle points} occur at any critical point $\bar{\mb U}$ that is not generated by choosing $r$ lead eigenvectors. Suppose that $\bar{\mb U} = \mb \Phi \mb \Gamma$, and $\mb \phi_\ell = \mb 0$ for some $\ell > r$. 
Then there is some lead eigenvector $\pm \mb \xi_i$ ($i \le r$) which does not participate in $\mb \Phi$. Consider a perturbation $\pm \mb W = \mb \xi_i \mb e_\ell^* \mb \Gamma$ which moves $\mb \phi_\ell$ in the direction of the neglected eigenvector $\mb \xi_i$ or its negative. The second derivative of $\varphi$ in this direction is simply
\begin{equation}
\nabla^2 \varphi [\mb W,\mb W] = \innerprod{ \mb U\mb U^* - \mb Y }{ \mb \xi_i \mb \xi_i^* } = - \lambda_i < 0.
\end{equation}
In words: the objective function exhibits strict negative curvature in any direction that perturbs any zero column $\mb \phi_\ell$ in the direction of a neglected. This is intuitive, since this modification allows the approximation to capture more of the energy of the observation $\mb Y$. Because we can perturb in either the direction of $+\mb \xi_\ell$ or $-\mb \xi_\ell$, this can be interpreted as negative curvature in a direction that breaks this symmetry. 

\item {\em Local minimizers} occur only at the global minimizers, of the form \eqref{eqn:opt-factorization}. There is a manifold of local minimizers, isometric to $O(r)$. This generalizes the ``circle'' of local minimizers observed in phase retrieval. 
\end{itemize}

\paragraph{General Low Rank Matrix Factorization}
In comparison to symmetric matrix factorization, the problem of factorizing a general rectangular matrix 
\begin{align}
\mb X=\mb U\mb V^*,\quad\mb U\neq\mb V
\end{align}
admits a full generalized linear symmetry. As mentioned above, this can be reduced to an orthogonal symmetry, by introducing an additional penalty, solving
\begin{align}
\min_{\mb U,\mb V}\quad \varphi(\mb U,\mb V)+\rho_s(\mb U,\mb V).
\end{align}
For example, setting $\rho_{s}(\mb U,\mb V) = \norm{\mb U^*\mb U-\mb V^*\mb V}F^2$ achieves this.\footnote{Other penalties are also possible -- e.g., $\rho_{s}(\mb U,\mb V)=\norm{\mb U}F^2+\norm{\mb V}F^2$, which encourages the factors to be balanced. This penalty is tightly connected to nuclear norm regularization.} 

How does this penalized problem behave? The critical points of this more general model also admit a simple description in terms of the spectral structure of the target matrix $\mb X_0$, given by the singular value decomposition $\mb X_0 = \sum_i \sigma_i \mb \xi_i \mb \nu_i^*$. Where in the symmetric case, critical points are generated by selecting subsets of eigenvectors, here every critical point $(\mb U,\mb V)$ is generated by selecting subsets of {\em singular} vectors:
\begin{eqnarray}
\mb U &=& \left[ \mb \phi_1 \mid \dots \mid \mb \phi_r \right] \mb \Gamma \\
\mb V &=& \left[ \mb \zeta_1 \mid \dots \mid \mb \zeta_r \right] \mb \Gamma, \qquad \mb \Gamma \in O(r),
\end{eqnarray}
where
\begin{eqnarray}
\mb \phi_1 &=& \sigma_{i_1}^{1/2} \mb \xi_{i_1}, \dots, \mb \phi_{r'} = \sigma_{i_{r'}}^{1/2} \mb \xi_{i_{r'}}, \\
\mb \zeta_1 &=& \sigma_{i_1}^{1/2} \mb \nu_{i_1}, \dots, \mb \zeta_{r'} = \sigma_{i_{r'}}^{1/2} \mb \nu_{i_{r'}}, 
\end{eqnarray}
and any remaining columns $\mb \phi_\ell, \mb \zeta_\ell$ are zero. This generalizes in a straightforward way the characterization for symmetric matrices above. Similar considerations show negative curvature in directions $\pm \left( \mb \xi_i \mb e_\ell^*, \mb \nu_i \mb e_\ell^* \right)$ that swap in a leading singular vector pair.


\section{Dispersive Structure: Negative Curvature}
\label{app:dispersive}

\begin{figure}[t]
\centerline{
\begin{tikzpicture}
\node at (0,0) {\includegraphics[width=3in]{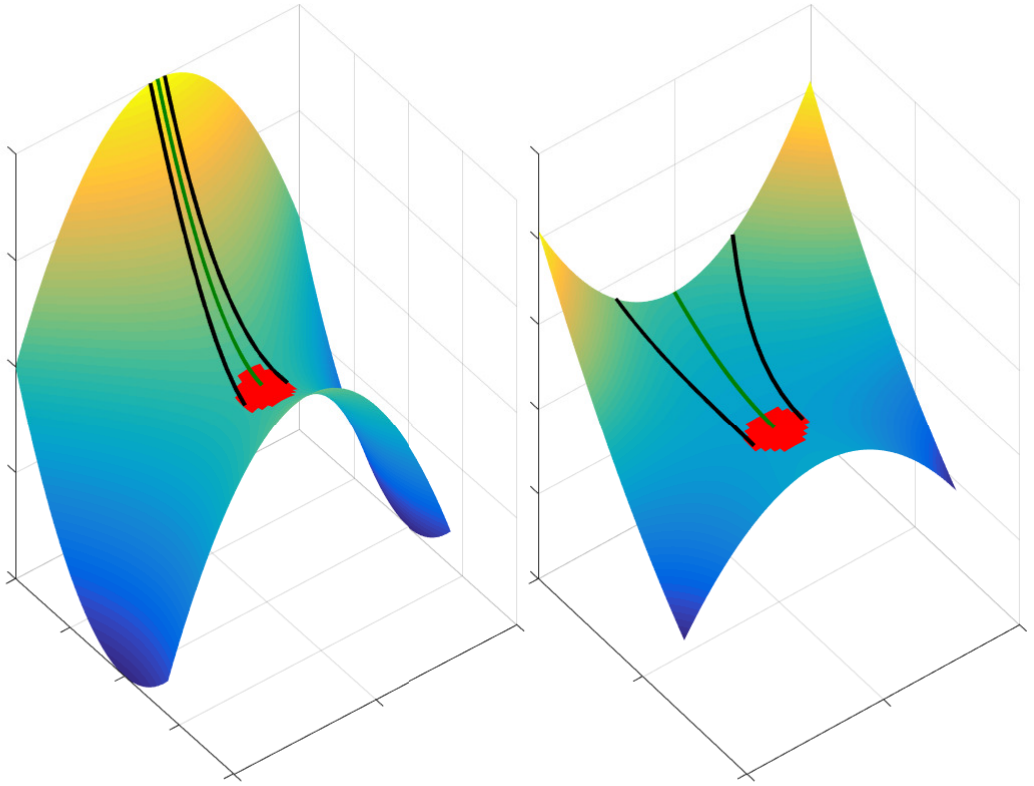}};
\node at (-2,-3.5) {\bf Dispersive};
\node at (2,-3.5) {\bf Non-Dispersive};
\end{tikzpicture}
}
\caption{{\bf Dispersive and Non-dispersive Flows}. Left: Dispersive functions exhibit global negative curvature. Right: nondispersive functions may exhibit positive curvature upstream of a saddle. In this situation, randomly initialized gradient descent may stagnate near the saddle point.}
\end{figure}

In this appendix, we describe in more detail the notion of {\em dispersion} illustrated in Figures \ref{fig:gpr}, \ref{fig:mf} and \ref{fig:huber-sphere} and the associated critical point diagrams. This notion seems to be important for explaining why the symmetric functions encountered in this paper are amenable to simple iterative methods such as randomly initialized gradient descent. The key intuition is that these functions exhibit a global negative curvature structure. This structure can be described most cleanly in terms of a continuous-time gradient flow 
\begin{equation}
\dot{\mb x}_t = - \nabla \varphi(\mb x_t). 
\end{equation}
We say that two critical points $\mb x^{\uparrow}$ and $\mb x^{\downarrow}$ are linked by gradient flow if there is a unique integral curve $\mb x_t$ of this differential equation, with $\lim_{t \to \infty} \mb x_t = \mb x^{\downarrow}$ and $\lim_{t \to -\infty} \mb x_t = \mb x^{\uparrow}$. 

We are interested in understanding whether gradient descent tends to stagnate near $\mb x^{\downarrow}$. To this end, it is useful to study the effect of perturbations $\mb v_t$ about $\mb x_t$. In particular, we are interested in understanding the behavior of the integral curve passing through $\mb x_t + \eps \mb v_t$, for $\eps$ small. Under gradient flow the perturbation evolves according to the linear time varying dynamical system 
\begin{equation} \label{eqn:perturbation}
\dot{\mb v}_t = - \nabla^2 \varphi(\mb x_t) \mb v_t.
\end{equation}
This differential equation provides a means of tracking the effect of perturbations across time. For any $t,t' \in (-\infty,\infty)$, we can define a transport operator $T_{t,t'}$ from the tangent space at $\mb x_t$ to the tangent space at $\mb x_{t'}$, by letting $T_{t',t} \mb w$ be the unique solution $\mb v_{t'}$ to the differential equation \eqref{eqn:perturbation} with initial condition $\mb v_t = \mb w$. 

The geometry of $\varphi$ around the saddle point $\mb x^\downarrow$ can be studied through the Hessian $\nabla^2 \varphi(\mb x^\downarrow)$. We say that the function $\varphi$ is {\em dispersive} along the path $\mb x^\uparrow \to \mb x^\downarrow$ if for every eigenvector $\mb v$ of $\nabla^2 \varphi(\mb x^\downarrow)$ that corresponds to a negative eigenvalue, and every $t_{\uparrow} \in (-\infty,\infty)$, there exists a $t_{\downarrow}$ such that for every $t_{\uparrow} \le t' \le t_{\downarrow} < t$, $\mb w = T_{t',t} \Pi_{\mb x_t,\mb x^\downarrow} \mb v$ is a direction of negative curvature, i.e., $\mb w^* \nabla^2 \varphi(\mb x_{t'}) \mb w < 0$.

This somewhat cumbersome technical definition exists to capture the idea that {\em downstream negative curvature directions are the images of upstream negative curvature directions under gradient flow}. All of the symmetric functions studied in this paper exhibit this property. However, worst case strict saddle functions such as the ``octopus function'' \cite{du2017gradient} do not. Intuitively speaking, this negative curvature structure helps gradient descent to avoid stagnating near saddle points. This intuition has been made formal in a number of special cases: generalized phase retrieval, complete dictionary learning, and multichannel deconvolution, etc.

\end{document}